\documentclass[11pt]{article}

\title{Improving the stability of the covariance-controlled adaptive Langevin thermostat for large-scale Bayesian sampling}

\author{Jiani Wei and Xiaocheng Shang\footnote{Corresponding author. Email: \href{mailto:x.shang.1@bham.ac.uk}{x.shang.1@bham.ac.uk} } \\
\small{School of Mathematics, University of Birmingham, Edgbaston, Birmingham, B15 2TT, United Kingdom} }

\date{\today}

\usepackage{cite}

\usepackage{hyperref}
\usepackage{graphicx}

\usepackage{caption}
\DeclareGraphicsExtensions{.eps,.mps,.pdf,.jpg,.png}
\graphicspath{{figures/}{../figures/}}

\usepackage{latexsym}
\usepackage{mathrsfs}
\usepackage{amssymb}
\usepackage{multirow}
\usepackage{booktabs}

\usepackage{fancyhdr}

\usepackage{amsmath}
\usepackage{bm}

\usepackage{amsfonts}

\usepackage{dsfont}

\usepackage[titletoc,toc,title]{appendix}

\usepackage[margin=3.0cm]{geometry}

\newcommand{\dd}{{\rm d}}

\newcommand{\p}{{\mathbf{p}}}
\newcommand{\m}{{\mathbf{m}}}
\newcommand{\x}{{\mathbf{x}}}
\newcommand{\F}{{\mathbf{F}}}
\newcommand{\I}{{\mathbf{I}}}
\newcommand{\M}{{\mathbf{M}}}
\newcommand{\V}{{\mathbf{V}}}

\newcommand{\Ebb}{{\mathbb{E}}}
\newcommand{\Ncal}{{\mathcal{N}}}
\newcommand{\thetaB}{{\boldsymbol{\theta}}}

\usepackage{mathrsfs}

\usepackage{soul}
\usepackage{color}

\usepackage{algorithm}
\usepackage{algorithmic}

\usepackage{url}
\usepackage{subfigure}

\usepackage{epstopdf}
\DeclareGraphicsExtensions{.eps,.mps,.pdf,.jpg,.png}

\usepackage{latexsym}
\usepackage{mathrsfs}
\usepackage{amssymb}
\usepackage{enumitem}

\usepackage{url}

\usepackage{array}
\newcolumntype{C}[1]{>{\centering\let\newline\\\arraybackslash\hspace{0pt}}m{#1}}

\begin{document}

\maketitle

\begin{abstract}
  Stochastic gradient Langevin dynamics and its variants approximate the likelihood of an entire dataset, via random (and typically much smaller) subsets, in the setting of Bayesian sampling. Due to the (often substantial) improvement of the computational efficiency, they have been widely used in large-scale machine learning applications. It has been demonstrated that the so-called covariance-controlled adaptive Langevin (CCAdL) thermostat, which incorporates an additional term involving the covariance matrix of the noisy force, outperforms popular alternative methods. A moving average is used in CCAdL to estimate the covariance matrix of the noisy force, in which case the covariance matrix will converge to a constant matrix in long-time limit. Moreover, it appears in our numerical experiments that the use of a moving average could reduce the stability of the numerical integrators, thereby limiting the largest usable stepsize. In this article, we propose a modified CCAdL (i.e., mCCAdL) thermostat that uses the scaling part of the scaling and squaring method together with a truncated Taylor series approximation to the exponential to numerically approximate the exact solution to the subsystem involving the additional term proposed in CCAdL. We also propose a symmetric splitting method for mCCAdL, instead of an Euler-type discretisation used in the original CCAdL thermostat. We demonstrate in our numerical experiments that the newly proposed mCCAdL thermostat achieves a substantial improvement in the numerical stability over the original CCAdL thermostat, while significantly outperforming popular alternative stochastic gradient methods in terms of the numerical accuracy for large-scale machine learning applications.
\end{abstract}

\pagenumbering{arabic}

\section{Introduction}
\label{sec:Introduction}

In many large-scale machine learning applications, it is computationally infeasible to directly generate samples with an entire dataset, for example, in standard Markov chain Monte Carlo (MCMC) methods~\cite{Metropolis1953} and hybrid Monte Carlo (HMC) methods~\cite{Brooks2011,Duane1987,Horowitz1991}.
In order to improve the computational efficiency, a number of the so-called stochastic gradient MCMC methods~\cite{Chen2014,Ding2014,Kim2025,Ma2015,Ma2017,Nemeth2021,Shang2015,Vollmer2015,Welling2011,Wu2025a} have been proposed in Bayesian sampling. These methods approximate the likelihood of the entire dataset, via random (and typically much smaller) subsets, thereby substantially reducing the computational cost in practice.

\subsection{Stochastic gradient}

In a typical setting of Bayesian sampling~\cite{Brooks2011,Robert2004}, we are interested in sampling from a posterior distribution defined as
\begin{equation}\label{eq:posterior_Bayesian}
  \pi(\thetaB|\mathbf{X}) \propto \pi(\mathbf{X}|\thetaB) \pi(\thetaB) \, ,
\end{equation}
where $\thetaB \in \mathbb{R}^{N_{\mathrm{d}}}$ denotes the parameter vector of interest, $\mathbf{X}$ is the entire dataset, and $\pi(\mathbf{X} | \thetaB)$ and $\pi(\thetaB)$ respectively represent the likelihood and prior distributions. As in molecular dynamics, a potential energy function $U(\thetaB)$ is introduced by setting $\pi(\thetaB|\mathbf{X}) \propto \exp(-\beta U(\thetaB))$, where $\beta$ is a positive parameter that is proportional to the reciprocal temperature in an associated physical system, i.e., $\beta^{-1}=k_{\mathrm{B}}T$, with $k_{\mathrm{B}}$ being the Boltzmann constant and $T$ the temperature. 
By taking the logarithm of~\eqref{eq:posterior_Bayesian}, we can rewrite the potential energy function as
\begin{equation}
  U(\thetaB) = -\log \pi(\mathbf{X}|\thetaB) - \log \pi(\thetaB) \, .
\end{equation}
Based on the assumption that the data points are independent and identically distributed (i.i.d.), the negative gradient of the potential energy can be written as
\begin{equation}\label{eq:gradpotential_energy}
  - \nabla U(\thetaB) = \sum_{i=1}^N  \nabla \log \pi(\x_{i}|\thetaB) + \nabla \log \pi(\thetaB) \, , 
\end{equation}
where $N$ denotes the size of the entire dataset. However, in many large-scale machine learning applications with a very large $N$, evaluations based on the entire dataset at each iteration, for example in MCMC and HMC methods, are computationally infeasible. In order to reduce the computational cost, random (and typically much smaller) subsets (also known as ``subsampling'') are preferred in stochastic gradient methods, in which case the log likelihood of the dataset is approximated as
\begin{equation}\label{eq:subsampling}
  \log \pi(\mathbf{X}|\thetaB) \approx \frac{N}{n}\sum_{i=1}^{n}\log \pi(\x_{r_{i}}|\thetaB) \, ,
\end{equation}
where $\{\x_{r_i}\}_{i=1}^n$ is a subset of the entire dataset $\mathbf{X}$, randomly selected at each iteration, and $n \ll N$ denotes the size of the subset. In this case,~\eqref{eq:gradpotential_energy} becomes a ``noisy force'' $\tilde{\F}(\thetaB) \in \mathbb{R}^{N_{\mathrm{d}}}$, which can be written as
\begin{equation}\label{eq:noisygradpotential_energy}
  \tilde{\F}(\thetaB) = - \nabla \tilde{U}(\thetaB) =  \frac{N}{n} \sum_{i=1}^n \nabla \log \pi(\x_{r_{i}}|\thetaB) + \nabla \log \pi(\thetaB) \, .
\end{equation}
As in~\cite{Chen2014,Ding2014,Shang2015}, we assume the stochastic gradient noise to be Gaussian with mean zero and unknown covariance, in which case the noisy force~\eqref{eq:noisygradpotential_energy} can be written as
\begin{equation}\label{eq:noisy_force}
  \tilde{\F}(\thetaB) = - \nabla U(\thetaB) + \sqrt{\boldsymbol{\Sigma}(\thetaB)} \M^{1/2} \mathbf{R} \, ,
\end{equation}
where $\boldsymbol{\Sigma}(\thetaB) \in \mathbb{R}^{N_{\mathrm{d}} \times N_{\mathrm{d}}}$ represents the covariance matrix of the noisy force, $\M \in \mathbb{R}^{N_{\mathrm{d}} \times N_{\mathrm{d}}}$ is typically a diagonal mass matrix, and $\mathbf{R} \in \mathbb{R}^{N_{\mathrm{d}}}$ denotes a vector of i.i.d.\ standard normal random variables. 
As explained in~\cite{Leimkuhler2015a}, in a typical setting of numerical integration with an associated stepsize $h$, we have
\begin{equation}
  h \tilde{\F}(\thetaB) = h \left( - \nabla U(\thetaB) + \sqrt{\boldsymbol{\Sigma}(\thetaB)} \M^{1/2} \mathbf{R} \right) = - h \nabla U(\thetaB) + \sqrt{h} \left( \sqrt{ h \boldsymbol{\Sigma}(\thetaB) } \right) \M^{1/2} \mathbf{R} \, .
\end{equation}

\subsection{Related work} 

Proposed by Welling and Teh, the stochastic gradient Langevin dynamics (SGLD)~\cite{Welling2011} combines the ideas of stochastic optimisation~\cite{Robbins1951} and traditional Brownian dynamics, with a sequence of stepsizes decreasing to zero. However, a fixed stepsize is often preferred in practice, which we also adopt in this article, as in Vollmer et al.~\cite{Vollmer2015}, where a modified SGLD (mSGLD) method was proposed in order to reduce the sampling bias.

Popular alternative methods include the stochastic gradient Hamiltonian Monte Carlo (SGHMC) method proposed by Chen et al.~\cite{Chen2014} and the stochastic gradient Nos\'{e}--Hoover thermostat (SGNHT) proposed by Ding et al.~\cite{Ding2014}, both of which will be included in numerical comparisons in this article. Based on a second order Langevin dynamics, SGHMC includes a parameter-dependent diffusion matrix in order to effectively offset the stochastic perturbation of the gradient. As a result, SGHMC is able to maintain a desired stationary distribution. In contrast, it has been demonstrated in~\cite{Leimkuhler2015a} that first order dynamics, such as SGLD (with a fixed stepsize), is unable to dissipate excess noise in gradient approximations, in which case first order dynamics is unlikely to be able to maintain a desired stationary distribution. However, it turns out that, the additional diffusion matrix in SGHMC is difficult to accommodate in practice, which may lead to a significant adverse effect on the sampling of the target distribution~\cite{Ding2014}. SGNHT adopts the ``thermostat'' idea that is widely used in molecular dynamics~\cite{Frenkel2001,Leimkuhler2015b}. It adjusts the kinetic energy during the simulation in order to maintain a constant (average) temperature, leading to the preservation of the canonical ensemble. SGNHT is also able to dissipate excess noise in gradient approximations and also maintains a desired stationary distribution~\cite{Leimkuhler2015a}. However, it relies on the assumption that the covariance matrix of the noisy force is constant, which may not be the case in practice.

\subsection{Covariance-controlled adaptive Langevin thermostat} 

It is worth mentioning that, in SGNHT, the covariance matrix is assumed to be constant, i.e., $\boldsymbol{\Sigma}=\sigma^{2}\I$, where $\sigma$ is a positive constant and $\I \in \mathbb{R}^{N_{\mathrm{d}} \times N_{\mathrm{d}}}$ is the identity matrix.
However, as explained in~\cite{Shang2015}, it is essential to consider parameter-dependent noise as the covariance matrix $\boldsymbol{\Sigma}(\thetaB)$ is expected to be naturally dependent on the parameter. 
To this end, proposed by Shang et al.~\cite{Shang2015}, the CCAdL thermostat generalises the idea of SGNHT and, as in SGHMC, also incorporates an additional term involving the covariance matrix of the noisy force, which is intended to effectively offset the stochastic perturbation of the gradient. CCAdL can be written as a standard It\={o} stochastic differential equation (SDE) system:
\begin{subequations}\label{eq:Ad-L-3}
  \begin{align}
    \dd \thetaB &= \M^{-1} \p \dd t \, , \\
    \dd \p &= -\nabla U(\thetaB)\dd t + \sqrt{ h \boldsymbol{\Sigma}(\thetaB) } \M^{1/2} \dd \mathbf{W} - (h/2) \beta \boldsymbol{\Sigma}(\thetaB) \p\dd t - \xi \p\dd t + \sqrt{2 A \beta^{-1}}\M^{1/2} \dd \mathbf{W}_{\mathrm{A}} \, , \\
    \dd \xi &= {\mu}^{-1} \left[ \p^{\mathsf{T}}\M^{-1}\p - N_{\mathrm{d}}k_{\mathrm{B}}T \right] \dd t \, ,
  \end{align}
\end{subequations}
where $\p \in \mathbb{R}^{N_{\mathrm{d}}}$ represents the associated momenta, $\xi \in \mathbb{R}$ is an auxiliary variable governed by a Nos\'{e}--Hoover device~\cite{Hoover1991,Nose1984a} via a negative feedback mechanism (see more discussions in the original article~\cite{Shang2015}), 
$\mu$ is a coupling parameter that is referred to as the ``thermal mass'' in the molecular dynamics setting, and $\dd {\mathrm{{\bf W}}}$ and $\dd {\mathrm{{\bf W}}_{\rm A}}$ respectively represent vectors of independent Wiener increments (often informally denoted by $\mathcal{N}(\mathbf{0}, \dd t \I)$~\cite{Chen2014}). Note that the coefficient $\sqrt{2 A \beta^{-1}}\M^{1/2} $ represents the strength of the artificial noise added into the system in order to improve the ergodicity, where the ``effective friction'' $A$ is a positive parameter that is proportional to the variance of the artificial noise. The incorporation of the additional term, $- (h/2)\beta \boldsymbol{\Sigma}(\thetaB) \p\dd t$, involving the parameter-dependent covariance matrix, is intended to offset the parameter-dependent noise also involving the covariance matrix~\eqref{eq:noisy_force}. As a result, it can be shown that the CCAdL thermostat~\eqref{eq:Ad-L-3} preserves a modified Gibbs stationary distribution~\cite{Shang2015}:
\begin{equation}\label{eq:Invariant_Dist_Modified_2}
  \hat{\rho}_{\beta}(\thetaB,\p,\xi) = Z^{-1}\exp\left({-\beta H(\thetaB,\p)}\right)\exp\left( - \beta\mu (\xi - A)^{2}/2 \right) \, .
\end{equation}
where $Z$ is the normalising constant and $H(\thetaB,\p) = \p^{\mathsf{T}}\M^{-1}\p/2 + U(\thetaB)$ is the Hamiltonian.

In practice, we do not know $\boldsymbol{\Sigma}(\thetaB)$ a priori. Therefore, we have to estimate $\boldsymbol{\Sigma}(\thetaB)$ during the simulation.
In the original CCAdL article~\cite{Shang2015}, we have $\boldsymbol{\Sigma}(\thetaB_t) = N^2 \I_t /n$ and $g(\thetaB; \x) = \nabla_{\thetaB} \log \pi(\x|\thetaB)$. It was assumed that $\thetaB_t$ does not change dramatically over time and a moving average~\cite{Ahn2012} was used to estimate $\I_t$:
\begin{equation}\label{eq:moving_averaging}
  \hat{\I}_t = (1-\kappa_t)\hat{\I}_{t-1} + \kappa_t \V(\thetaB_t) \, , 
\end{equation}
where $\kappa_t = 1/t$ and
\begin{equation}
  \V(\thetaB_t) = \frac{1}{n-1}\sum_{i=1}^n \left(g(\thetaB_t;\x_{r_{i}}) - \bar{g}(\thetaB_t) \right) \left(g(\thetaB_t;\x_{r_{i}}) - \bar{g}(\thetaB_t) \right)^{\mathsf{T}}
\end{equation}
is the empirical covariance of the gradient, with $\bar{g}(\thetaB_t)$ being the mean gradient of the log likelihood computed from a subset
\begin{equation}
  \bar g(\thetaB_t) = \frac{1}{n}\sum_{i=1}^n g(\thetaB_t;\x_{r_i}) \, .
\end{equation}
One potential issue of the moving average~\eqref{eq:moving_averaging} is that the covariance matrix will converge to a constant matrix in long-time limit (i.e., $\hat{\I}_t \to \hat{\I}_{t-1}$ as $t \to \infty$ in~\eqref{eq:moving_averaging}). However, the covariance matrix is naturally parameter-dependent. Moreover, it is computationally infeasible to estimate the full covariance matrix in high dimension, for which a diagonal approximation of the covariance matrix was employed in high dimension in the original CCAdL article~\cite{Shang2015}. Although the diagonal approximation of the covariance matrix worked well in some of the numerical experiments in high dimension in~\cite{Shang2015}, we could numerically integrate the subsystem involving the additional term without estimating the full covariance matrix in order to reduce the computational cost. 

\subsection{Contributions} 

In this article, we propose a modified CCAdL (mCCAdL) thermostat that does not rely on the moving average. Moreover, mCCAdL does not need to explicitly estimate the full covariance matrix in integrating the subsystem involving the additional term; instead, it makes use of the fact that the exact solution to the subsystem involving the additional term is essentially a product of a matrix exponential and a vector, which can be solved more efficiently in practice. We also propose a symmetric splitting method for mCCAdL, instead of an Euler-type discretisation used in the original CCAdL thermostat. Our numerical experiments in large-scale machine learning applications demonstrate that the newly proposed mCCAdL thermostat substantially outperforms popular alternative methods in terms of the numerical stability (measured by the largest usable stepsize) and accuracy (measured by the test error), leading to significantly improved computational efficiency. 

The rest of the article is organised as follows. 
In Section~\ref{sec:mCCAdL}, we propose the mCCAdL thermostat that uses a novel approach to integrate the subsystem involving the additional term; we also propose a symmetric splitting method for the mCCAdL thermostat. Various numerical experiments are reported in Section~\ref{sec:Numerical_Experiments} to evaluate the performance of mCCAdL in large-scale machine learning applications. Finally, Section~\ref{sec:Conclusions} summarises our findings.

\section{Modified covariance-controlled adaptive Langevin thermostat}
\label{sec:mCCAdL}

In this section, we introduce the mCCAdL method, including the estimation of the parameter-dependent covariance matrix in Section~\ref{subsec:covariance_matrix}, the integration of the subsystem involving the additional term in Section~\ref{subsec:additional_term}, and the design of splitting methods in Section~\ref{subsec:splitting_methods}.

\subsection{Estimating the parameter-dependent covariance matrix}
\label{subsec:covariance_matrix}

In addition to the assumption that the stochastic gradient noise follows a normal distribution~\eqref{eq:noisy_force}, we further assume that the size of the random subset $n$ is large enough for the central limit theorem to hold, in which case we have
\begin{equation}
  \frac{1}{n} \sum_{i=1}^n g(\thetaB_t; \x_{r_{i}}) \sim \Ncal \left(\Ebb_{\x} [g(\thetaB_t; \x)], \frac{1}{n} \mathrm{Cov}[g(\thetaB_t; \x)] \right) \, ,
\end{equation}
where $\Ebb_{\x} [g(\thetaB_t; \x)]$ and $\mathrm{Cov}[g(\thetaB_t; \x)]$ are the expectation and covariance of the gradient $g(\thetaB_t; \x) = \nabla_{\thetaB_t} \log \pi(\x|\thetaB_t)$, respectively. We can easily see that the expectation of the noisy force~\eqref{eq:noisygradpotential_energy} is equal to that of the clean (full) force~\eqref{eq:gradpotential_energy}, that is, $\Ebb_{\x}[-\nabla \tilde{U}(\thetaB_t)] = \Ebb_{\x}[-\nabla U(\thetaB_t)] $, and therefore the noisy force can be written as
\begin{equation}
  - \nabla \tilde{U}(\thetaB_t) = - \nabla U(\thetaB_t) + \Ncal \left(\mathbf{0}, \boldsymbol{\Sigma}(\thetaB_t) \right) \, ,
\end{equation}
where the covariance matrix of the noisy force is calculated as
\begin{equation}\label{eq:covariance_matrix}
  \boldsymbol{\Sigma}(\thetaB_t) \equiv \frac{N^2}{n} \mathrm{Cov}[g(\thetaB_t; \x)] = \frac{N^2}{n} \V(\thetaB_t) \, .
\end{equation}
Note that we no longer use the moving average~\eqref{eq:moving_averaging} and instead estimate the covariance matrix ``directly'' as in~\eqref{eq:covariance_matrix}.

\subsection{Integrating the subsystem involving the additional term}
\label{subsec:additional_term}

Let us now consider the subsystem involving the additional term in~\eqref{eq:Ad-L-3}:
\begin{equation}\label{eq:additional_term}
  \dd \p = - (h/2) \beta \boldsymbol{\Sigma}(\thetaB) \p\dd t \, .
\end{equation}
It turns out that we do not need to estimate the full covariance matrix $\boldsymbol{\Sigma}(\thetaB)$ before numerically integrating the subsystem~\eqref{eq:additional_term} using for instance an Euler-type integrator as in the original CCAdL article~\cite{Shang2015}. Instead, we can make use of the fact that there exists an exact solution to~\eqref{eq:additional_term}:
\begin{equation}\label{eq:additional_term_sols}
  \p(t) = e^{t \tilde{\boldsymbol{\Sigma}}(\thetaB)} \p(0) \, ,
\end{equation}
where $\tilde{\boldsymbol{\Sigma}}(\thetaB) = -(h/2) \beta \boldsymbol{\Sigma}(\thetaB) \in \mathbb{R}^{N_{\mathrm{d}} \times N_{\mathrm{d}}}$ is a symmetric matrix. Subsequently, following~\cite{Al-Mohy2011}, we can use the scaling part of the scaling and squaring method together with a truncated Taylor series approximation to the exponential to numerically approximate the exact solution in~\eqref{eq:additional_term_sols}, which is essentially a product of a matrix exponential and a vector.

In order to reduce the computational cost, we first shift the matrix $\tilde{\boldsymbol{\Sigma}}$ to obtain $\hat{\boldsymbol{\Sigma}} = \tilde{\boldsymbol{\Sigma}} - \tilde{\mu} \I$, where $\tilde{\mu} = \mathrm{tr}(\tilde{\boldsymbol{\Sigma}}) / N_{\mathrm{d}}$ and $\I$ is again the identity matrix. The shifting allows us to rewrite the exact solution in~\eqref{eq:additional_term_sols} as
\begin{equation}
  e^{t \tilde{\boldsymbol{\Sigma}}} \p(0) = e^{t (\hat{\boldsymbol{\Sigma}} + \tilde{\mu} \mathbf{I})} \p(0) = e^{t \tilde{\mu}} e^{t \hat{\boldsymbol{\Sigma}}} \p(0) \, .
\end{equation}
We further introduce a scaling parameter $s \geq 1$, which is an integer. This allows us to ``scale'' the exact solution in~\eqref{eq:additional_term_sols} as
\begin{equation}
  e^{t \tilde{\boldsymbol{\Sigma}}} \p(0) = e^{t \tilde{\mu}} \left( e^{\frac{t}{s} \hat{\boldsymbol{\Sigma}}} \right)^s \p(0) \, ,
\end{equation}
where each scaled matrix exponential $e^{\frac{t}{s} \hat{\boldsymbol{\Sigma}}}$ is approximated by using a truncated Taylor series of degree $m$:
\begin{equation}\label{eq:truncated_Taylor}
  e^{\frac{t}{s} \hat{\boldsymbol{\Sigma}}} \approx \sum_{j=0}^{m} \frac{1}{j!} \left( \frac{t}{s} \hat{\boldsymbol{\Sigma}} \right)^j \, .
\end{equation}
Subsequently, starting from $v_0 = \p(0)$, we iteratively compute
\begin{equation}
  v_{k+1} = e^{\frac{t}{s} \hat{\boldsymbol{\Sigma}}} v_k \, , \quad  k = 0, 1, \dots, s-1 \, .
\end{equation}
In the end, we multiply $v_s$ by $e^{t \tilde{\mu}}$ to rewrite the exact solution in~\eqref{eq:additional_term_sols} as
\begin{equation}\label{eq:scaling}
  e^{t \tilde{\boldsymbol{\Sigma}}} \p(0) = e^{t \tilde{\mu}} \underbrace{ \left(\sum_{j=0}^{m} \frac{1}{j!} \left( \frac{t}{s} \hat{\boldsymbol{\Sigma}} \right)^j \right) \cdot \left(\sum_{j=0}^{m} \frac{1}{j!} \left( \frac{t}{s} \hat{\boldsymbol{\Sigma}} \right)^j \right) \cdots \left(\sum_{j=0}^{m} \frac{1}{j!} \left( \frac{t}{s} \hat{\boldsymbol{\Sigma}} \right)^j \right) }_{s~\mathrm{times}} \p(0) \, . 
\end{equation}
Note that we can determine an optimal choice of $s$ and $m$ by exploiting the backward error analysis of Higham~\cite{Higham2005a,Higham2009}, as refined by Al-Mohy and Higham~\cite{Al-Mohy2010}. Moreover, one particular advantage of the current approach~\eqref{eq:scaling} is that we do not need to explicitly forming the matrix exponential, $\exp(t \tilde{\boldsymbol{\Sigma}})$, before computing its product with the vector, $\p(0)$. In what follows we refer the newly proposed approach as the mCCAdL thermostat.

\subsection{Splitting methods}
\label{subsec:splitting_methods}

The optimal design of splitting methods in ergodic SDEs has been an active area of research in the mathematics community~\cite{Leimkuhler2013c,Leimkuhler2015,Leimkuhler2015a,Leimkuhler2015b,Leimkuhler2016a,Shang2017,Shang2019,Shang2020,Duong2021,Wu2025}. It has been demonstrated in~\cite{Leimkuhler2015a} that one particular type of symmetric splitting methods for the SGNHT method (equivalent to the adaptive Langevin thermostat~\cite{Jones2011,Leimkuhler2015a}) with a clean (full) gradient (i.e., without subsampling) inherits the superconvergence property (i.e., fourth order convergence to a modified Gibbs stationary distribution for configurational quantities) recently proved in the setting of Langevin dynamics~\cite{Leimkuhler2013,Leimkuhler2013c}. However, a naive nonsymmetric splitting method was used for CCAdL in the original CCAdL article~\cite{Shang2015}. In what follows we propose a symmetric splitting method for mCCAdL. More specifically, the vector field of the CCAdL/mCCAdL system~\eqref{eq:Ad-L-3} is decomposed into five parts, which we label as ``A'', ``B'', ``C'', ``O'', and ``D'', respectively:
\begin{equation}\label{eq:Ad-L-3_Splitting}
\begin{aligned}
  \dd \left[ \begin{array}{c} \thetaB \\ \p \\ \xi \end{array} \right] = 
  & \, \underbrace{ \left[ \begin{array}{c} \M^{-1} \p \\ \mathbf{0} \\ 0 \end{array} \right]\dd t }_{\mathrm{A}} + \underbrace{ \left[ \begin{array}{c} \mathbf{0} \\ \tilde{\F}(\thetaB) \\ 0 \end{array} \right]\dd t }_{\mathrm{B}} + \underbrace{ \left[ \begin{array}{c} \mathbf{0} \\ - (h/2) \beta \boldsymbol{\Sigma}(\thetaB)\p \\ 0 \end{array} \right]\dd t }_{\mathrm{C}} \\
  & + \underbrace{ \left[ \begin{array}{c} \mathbf{0} \\ - \xi \p \dd t + \sqrt{2A\beta^{-1}} \M^{1/2} \dd \mathbf{W}_{\mathrm{A}} \\ 0 \end{array} \right] }_{\mathrm{O}} + \underbrace{ \left[ \begin{array}{c} \mathbf{0} \\ \mathbf{0} \\ {\mu}^{-1} \left[ \p^{\mathsf{T}}\M^{-1}\p - N_{\mathrm{d}}k_{\mathrm{B}}T \right] \end{array} \right]\dd t }_{\mathrm{D}} \, .
\end{aligned}
\end{equation}
As explained in detail in~\cite{Leimkuhler2015a}, parts ``A'', ``B'', ``O'', and ``D'' can all be solved ``exactly'', while part ``C'' can be approximated as described in the previous subsection. Note that part B is written slightly differently here as compared in~\eqref{eq:Ad-L-3}; this is to emphasise that in practice we do not need to explicitly estimate the covariance matrix appearing in part B---the noisy force $\tilde{\F}(\thetaB)$ is generated by the subsampling procedure~\eqref{eq:noisygradpotential_energy}. The generators associated with each part are defined, respectively, as
\begin{subequations}
\begin{align}
  \mathcal{L}_\mathrm{A} &= \M^{-1}\p \cdot \nabla_{\thetaB} \, , \\
  \mathcal{L}_\mathrm{B} &= - \nabla U(\thetaB) \cdot \nabla_{\p} + (h/2) \mathrm{tr} \left(\boldsymbol{\Sigma}(\thetaB)\M\nabla^{2}_{\p}\right) \, , \\
  \mathcal{L}_\mathrm{C} &= - (h/2) \beta \boldsymbol{\Sigma}(\thetaB) \p \cdot \nabla_{\p} \, , \\
  \mathcal{L}_\mathrm{O} &= - \xi \p \cdot \nabla_{\p} + A \beta^{-1} \mathrm{tr} \left(\M\nabla^{2}_{\p}\right) \, , \\
  \mathcal{L}_\mathrm{D} &= \mu^{-1} \left[ \p^{\mathsf{T}}\M^{-1}\p - N_{\mathrm{d}}k_{\mathrm{B}}T \right] \cdot \nabla_{\xi} \, .
\end{align}
\end{subequations}
Overall, the generator of the CCAdL/mCCAdL system~\eqref{eq:Ad-L-3} can be written
as
\begin{equation}
  \mathcal{L} = \mathcal{L}_\mathrm{A} + \mathcal{L}_\mathrm{B} + \mathcal{L}_\mathrm{C} + \mathcal{L}_\mathrm{O} + \mathcal{L}_\mathrm{D} \, .
\end{equation}
The flow map (or phase space propagator) of the system, $\mathcal{F}_{t} = \exp \left( t \mathcal{L} \right)$, is used to formally denote the solution operator. It turns out that a large number of approximations of $\mathcal{F}_{t}$ can be obtained as products (taken in different arrangements) of exponentials of the splitting terms. It has been demonstrated that different splittings and/or combinations give dramatically different performance in practice~\cite{Leimkuhler2015a}. It has also been well documented that symmetrical splitting methods often outperform their nonsymmetric counterparts (e.g., in the case of stochastic gradient methods~\cite{Chen2015,Leimkuhler2015a}). In this article, we propose the following symmetric splitting method, which we refer as the ``BAODCDOAB'' method:
\begin{equation}\label{eq:Propagator mCCAdL}
  e^{h\hat{\mathcal{L}}_{\mathrm{BAODCDOAB}}} = e^{\frac{h}{2}\mathcal{L}_\mathrm{B}} e^{\frac{h}{2}\mathcal{L}_\mathrm{A}} e^{\frac{h}{2}\mathcal{L}_\mathrm{O}} e^{\frac{h}{2}\mathcal{L}_\mathrm{D}} e^{h\hat{\mathcal{L}}_\mathrm{C}} e^{\frac{h}{2}\mathcal{L}_\mathrm{D}} e^{\frac{h}{2}\mathcal{L}_\mathrm{O}} e^{\frac{h}{2}\mathcal{L}_\mathrm{A}} e^{\frac{h}{2}\mathcal{L}_\mathrm{B}} \, ,
\end{equation}
where $\exp\left(h\mathcal{L}_f\right)$ denotes the phase space propagator related to the corresponding vector field $f$. Note that the steplengths associated with various operations are uniform, spanning the interval $h$. Therefore, each of the B, A, O, and D steps in~\eqref{eq:Propagator mCCAdL} is taken with a steplength of $h/2$, while a steplength of $h$ is used in the C step. Note that the C step only appears once in the sequence, meaning that we only need to approximate the associated C part of the system once at each iteration. The procedure of the BAODCDOAB method is summarised in Algorithm~\ref{alg:mCCAdL}.
It is worth mentioning that the noisy force computed at the end of each iteration can be reused at the start of the next iteration; therefore only one (noisy) force calculation is needed in the BAODCDOAB method at each iteration. Based on the Baker--Campbell--Hausdorff expansion (e.g., in~\cite{Leimkuhler2015a,Shang2020}), it can be demonstrated that the newly proposed symmetric splitting method associated with mCCAdL has a weak second order convergence to the modified Gibbs stationary distribution~\eqref{eq:Invariant_Dist_Modified_2}, if the approximation~\eqref{eq:scaling} in the C part is an at-least-second-order approximation to its exact solution~\eqref{eq:additional_term_sols}. Note that it is nontrivial to achieve a weak second order convergence to a stationary distribution when the stochastic gradient noise is present (even for symmetric splitting methods). In contrast, the methods associated with SGHMC, SGNHT, and CCAdL are all expected to have a weak first order convergence to their respective stationary distributions.

\begin{algorithm}[tb]
  \caption{The BAODCDOAB method for the mCCAdL thermostat.}
  \label{alg:mCCAdL}
\begin{algorithmic}[1]
  \STATE {\bfseries Input:} $\M$, $h$, $\beta$, $A$, $\mu$, and $N_{\mathrm{d}}$.
  \STATE {\bfseries Initialise } $\thetaB_{0}$, $\p_0$, and $\xi_0 = A$.
  \FOR{$t=1,2,\ldots, {\hat{T}}$}
  \STATE $\p_{t} \gets \p_{t-1} + (h/2) \tilde{\F}(\thetaB_{t-1})$;
  \STATE $\thetaB_{t} \gets \thetaB_{t-1} + (h/2) \M^{-1}\p_{t}$;
    \IF{$\xi_{t-1} \neq 0$}
      \STATE $\p_{t} \gets e^{-\xi_{t-1} (h/2)} \p_{t} + \sqrt{ A\beta^{-1} (1-e^{-\xi_{t-1} h}) / \xi_{t-1} } \M^{1/2} \mathcal{N}(\mathbf{0}, \I)$;
    \ELSE
      \STATE $\p_{t} \gets \p_{t} + \sqrt{hA\beta^{-1}} \M^{1/2} \mathcal{N}(\mathbf{0}, \I)$;
    \ENDIF
  \STATE $\xi_{t} \gets \xi_{t-1} + (h/2) {\mu}^{-1} \left[ \p^{\mathsf{T}}_{t}\M^{-1}\p_{t} - N_{\mathrm{d}}\beta^{-1} \right]$;
  \STATE Approximate the new $\p_{t}$ in~\eqref{eq:additional_term_sols} by using the \texttt{expmv} function in MATLAB;
  \STATE $\xi_{t} \gets \xi_{t} + (h/2) {\mu}^{-1} \left[ \p^{\mathsf{T}}_{t}\M^{-1}\p_{t} - N_{\mathrm{d}}\beta^{-1} \right]$;
    \IF{$\xi_{t} \neq 0$}
      \STATE $\p_{t} \gets e^{-\xi_{t} (h/2)} \p_{t} + \sqrt{ A\beta^{-1} (1-e^{-\xi_{t} h}) / \xi_{t} } \M^{1/2} \mathcal{N}(\mathbf{0}, \I)$;
    \ELSE
      \STATE $\p_{t} \gets \p_{t} + \sqrt{hA\beta^{-1}} \M^{1/2} \mathcal{N}(\mathbf{0}, \I)$;
    \ENDIF
  \STATE $\thetaB_{t} \gets \thetaB_{t} + (h/2) \M^{-1}\p_{t}$; 
  \STATE $\p_{t} \gets \p_{t} + (h/2) \tilde{\F}(\thetaB_{t})$;    
  \ENDFOR
\end{algorithmic}
\end{algorithm}

\section{Numerical experiments}
\label{sec:Numerical_Experiments}
In this section, we compare the newly established mCCAdL method with SGHMC~\cite{Chen2014}, SGNHT~\cite{Ding2014}, and CCAdL~\cite{Shang2015} on various machine learning tasks to demonstrate the benefits of mCCAdL in large-scale Bayesian sampling with a noisy force. In order to ensure a fair comparison, unless otherwise stated, we adopt the same experimental settings as in the original CCAdL article~\cite{Shang2015}, including the same set of parameters: $\M=\I$, $\beta=1$, and $\mu=N_{\mathrm{d}}$. The numerical experiments in Section~\ref{subsec:2W_distance} follow the settings in~\cite{Gurbuzbalaban2021} and were performed by using MATLAB R2024a on a macOS 14.5 system equipped with an Apple M1 processor (8 cores) and 8GB RAM. The numerical experiments in Section~\ref{subsec:binaryclass} were performed by using the same settings as in Section~\ref{subsec:2W_distance}. The numerical experiments in Section~\ref{subsec:DRBM} were performed by using MATLAB R2023b on a Linux server with Intel Xeon Platinum 8480CL CPUs (112 physical cores), where MATLAB was assigned 72 logical cores due to OS constraints. Note that GPU acceleration was not used throughout the article.

\subsection{2-Wasserstein distance evaluation with Bayesian linear regression}
\label{subsec:2W_distance}

We begin with a Bayesian linear regression problem and evaluate the samplers by considering the 2-Wasserstein distance between an empirical distribution and the true posterior distribution~\cite{Gurbuzbalaban2021}. We generated $N = 10{,}000$ data points with $N_{\mathrm{d}} = 100$ according to  
\begin{equation}
    \x_i \sim \mathcal{N}(\mathbf{0}, \I) \, , \quad y_i = \thetaB^{\mathsf{T}} \x_i + \delta_i \, ,
\end{equation}
where $\delta_i \sim \mathcal{N}(0, 1)$ are i.i.d.\ standard normal variables. We used a likelihood function of
\begin{equation}
  \pi\left( \mathbf{y}| \thetaB, \mathbf{X} \right) \propto \exp\!\left[ - \left( \mathbf{y} - \thetaB^{\mathsf{T}}\mathbf{X} \right)^{\mathsf{T}} \left( \mathbf{y} - \thetaB^{\mathsf{T}}\mathbf{X} \right) / 2 \right] \, ,
\end{equation}
where $\mathbf{X}=\left[\x_1,\ldots,\x_N\right] \in \mathbb{R}^{N_{\mathrm{d}} \times N}$ and $\mathbf{y}=\left[y_1,\ldots,y_N\right] \in \mathbb{R}^{1\times N}$, and a prior distribution of $\pi(\thetaB) \sim \Ncal(\mathbf{0}, \lambda \I)$, where we used $\lambda = 10$ in our numerical experiments. We then derived the posterior distribution as a Gaussian distribution $\pi(\thetaB | \mathbf{y}, \mathbf{X}) \sim \mathcal{N}(\m, \boldsymbol{\Sigma})$, where the mean vector is given by $\m = \boldsymbol{\Sigma} \mathbf{X} \mathbf{y}^{\mathsf{T}}$ with the covariance matrix being 
\begin{equation}
  \boldsymbol{\Sigma} = \left[ \mathbf{X} \mathbf{X}^{\mathsf{T}} + (\lambda \I)^{-1} \right]^{-1} \, .
\end{equation}
At each iteration $k$, we randomly selected a subset of size $n=500$ with replacement. We then formed an empirical Gaussian distribution $\mathcal{N}\left(\hat{\m}_k,\hat{\boldsymbol{\Sigma}}_k\right)$ and computed the 2-Wasserstein distance between the empirical distribution and the true posterior distribution, using an explicit formula that characterises the 2-Wasserstein
distance between any two Gaussian distributions~\cite{Givens1984}.
\begin{figure}[tb]
\begin{center}
\begin{tabular}{ccc}
\includegraphics[width=0.33\linewidth]{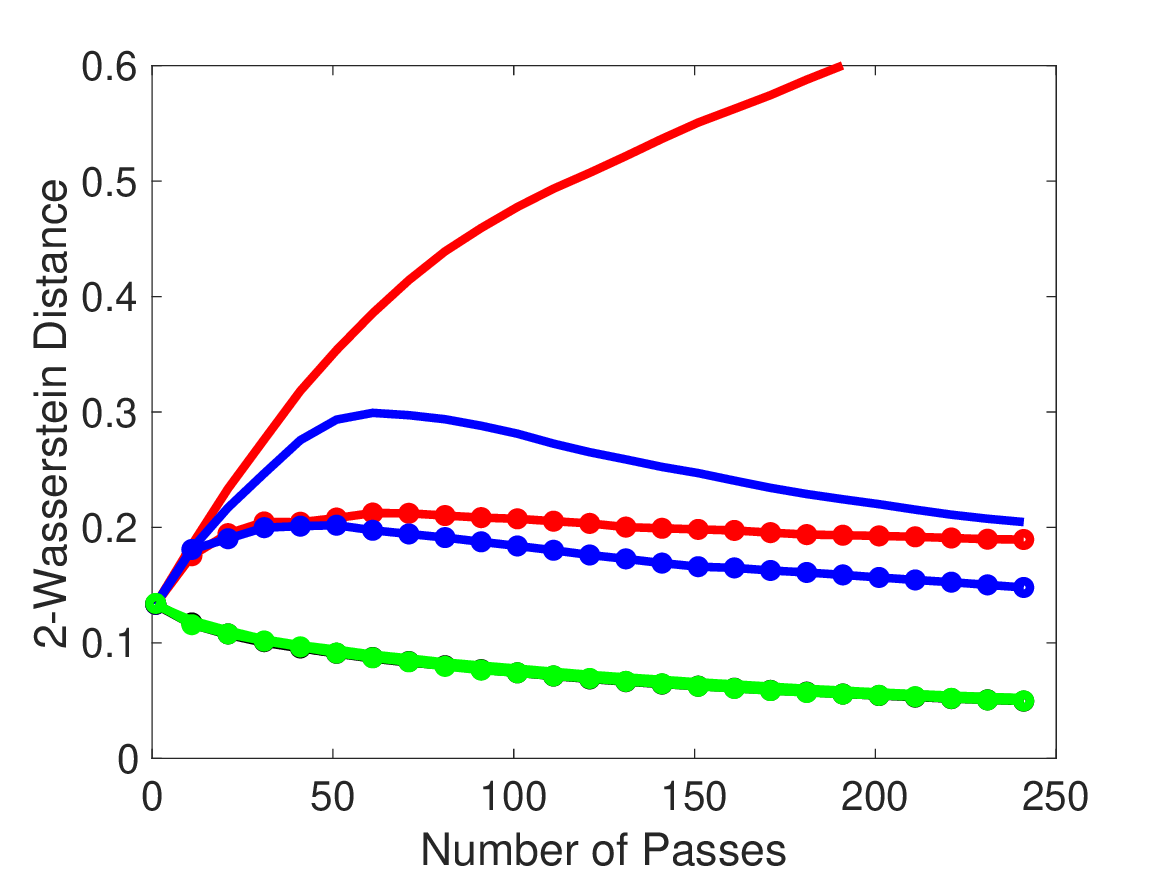} 
\hspace{-4.5mm}
&\includegraphics[width=0.33\linewidth]{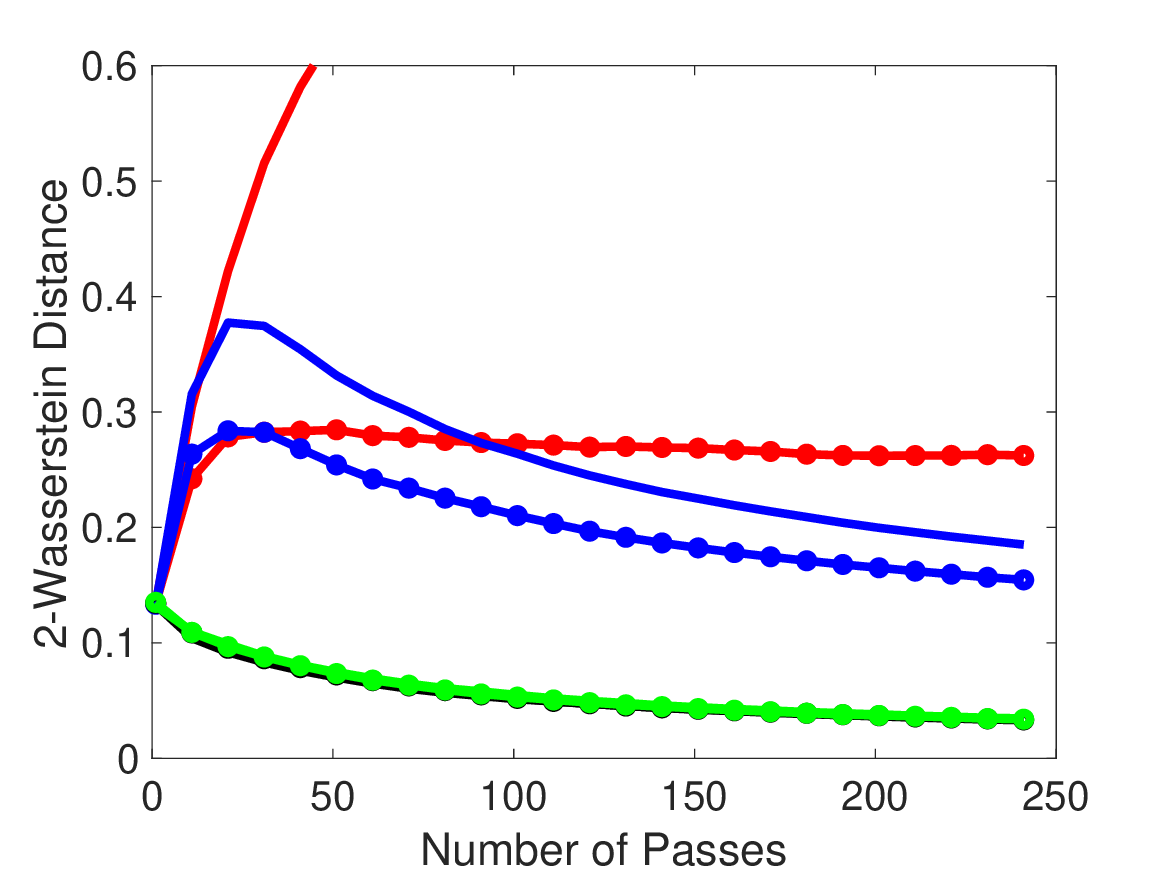} 
\hspace{-4.5mm}
&\includegraphics[width=0.33\linewidth]{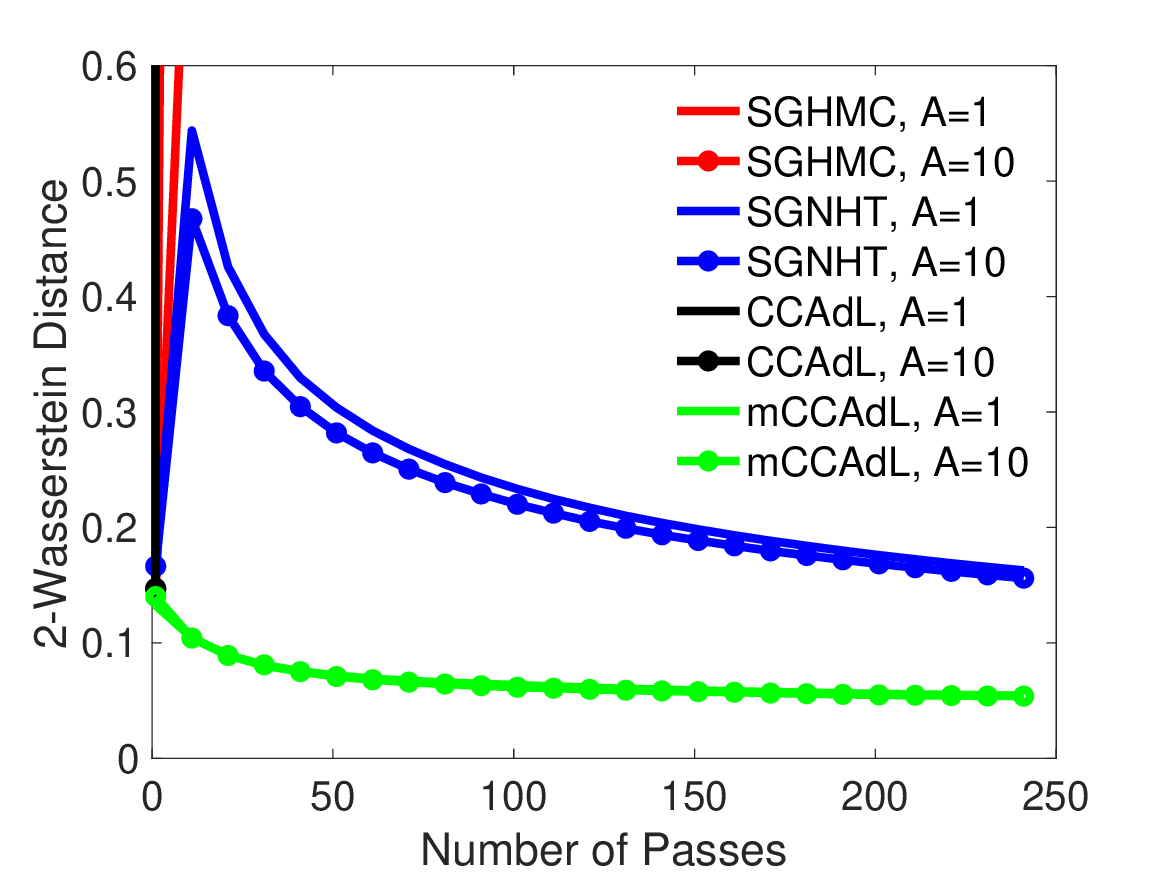}\\
\small{(a) $h=5 \times 10^{-4}$} \hspace{-5.5mm} & \small{(b) $h=1 \times 10^{-3}$} \hspace{-4.5mm} & \small{(c) $h=5 \times 10^{-3}$}
\end{tabular}
\end{center}
\caption{\small Comparisons of the 2-Wasserstein distance of various methods against the number of passes over the entire dataset in the Bayesian linear regression with various values of the stepsize $h$ and effective friction $A$.}
\label{fig:2W_distance}
\end{figure}
Figure~\ref{fig:2W_distance} reports the performance of various methods in terms of the 2-Wasserstein distance between the empirical distribution and the true posterior distribution, plotted against the number of passes through the entire dataset. We can observe that the newly established mCCAdL method achieves the smallest 2-Wasserstein distance with various values of the stepsize $h$, closely matching CCAdL when the stepsize $h$ is small. In contrast, both SGHMC and SGNHT display larger values of the 2-Wasserstein distance, although their performance improves with larger values of the effective friction $A$ (as can be seen on the left and middle panels). Moreover, as the stepsize increases, mCCAdL remains stable, while the other three alternative methods all becomes worse and worse, with both SGHMC and CCAdL blowing up when the stepsize is $h = 5 \times 10^{-3}$ on the right panel. 


\subsection{Binary classification with Bayesian logistic regression}
\label{subsec:binaryclass}

We consider a large-scale Bayesian logistic regression model trained on subsets of the benchmark Modified National Institute of Standards and Technology (MNIST) dataset\footnote{\small \url{http://yann.lecun.com/exdb/mnist/}} and Canadian Institute For Advanced Research (CIFAR-10) dataset\footnote{\small
\url{https://www.cs.toronto.edu/~kriz/cifar.html}}. We focus on the binary classification of digits 7 and 9 for the MNIST dataset, while three binary classification tasks, corresponding to the class pairs of airplane--automobile, deer--horse, and cat--dog, are considered for the CIFAR-10 dataset.

For the MNIST task, the training set consists of 12,214 samples, and the test set contains 2,037 samples. As in the original CCAdL article~\cite{Shang2015}, we applied a 100-dimensional random projection of the original features in order to reduce the dimensionality. For each CIFAR-10 binary classification task, the size of the training set is 10,000, and the size of the test set is 2,000. Unlike the MNIST task, the input features were projected onto a 400-dimensional space using a principal component analysis for dimensionality reduction as well as improving the convergence. We used a likelihood function of
\begin{equation}\label{eq:likelihood-MNIST}
  \pi\left(\mathbf{y}, \mathbf{X} | \thetaB \right) = \pi\left(\{y_i, \x_i\}_{i=1}^N | \thetaB \right) \propto \prod_{i =1}^N \frac{1}{ 1 + \exp\left(-y_i \thetaB^{\mathsf{T}} \x_i\right) } \, ,
\end{equation}
where the labels are assigned as either $y_i = 1$ (e.g., for images of digit 7) or $y_i = -1$ (e.g., for images of digit 9), and a Gaussian prior distribution of $\pi(\thetaB) \propto \exp\left( - \thetaB^{\mathsf{T}} \thetaB / 2 \right)$.
At each iteration, we randomly selected a subset of size $n = 500$ with replacement. 
\begin{figure}[tb]
\begin{center}
\begin{tabular}{ccc}
\includegraphics[width=0.33\linewidth]{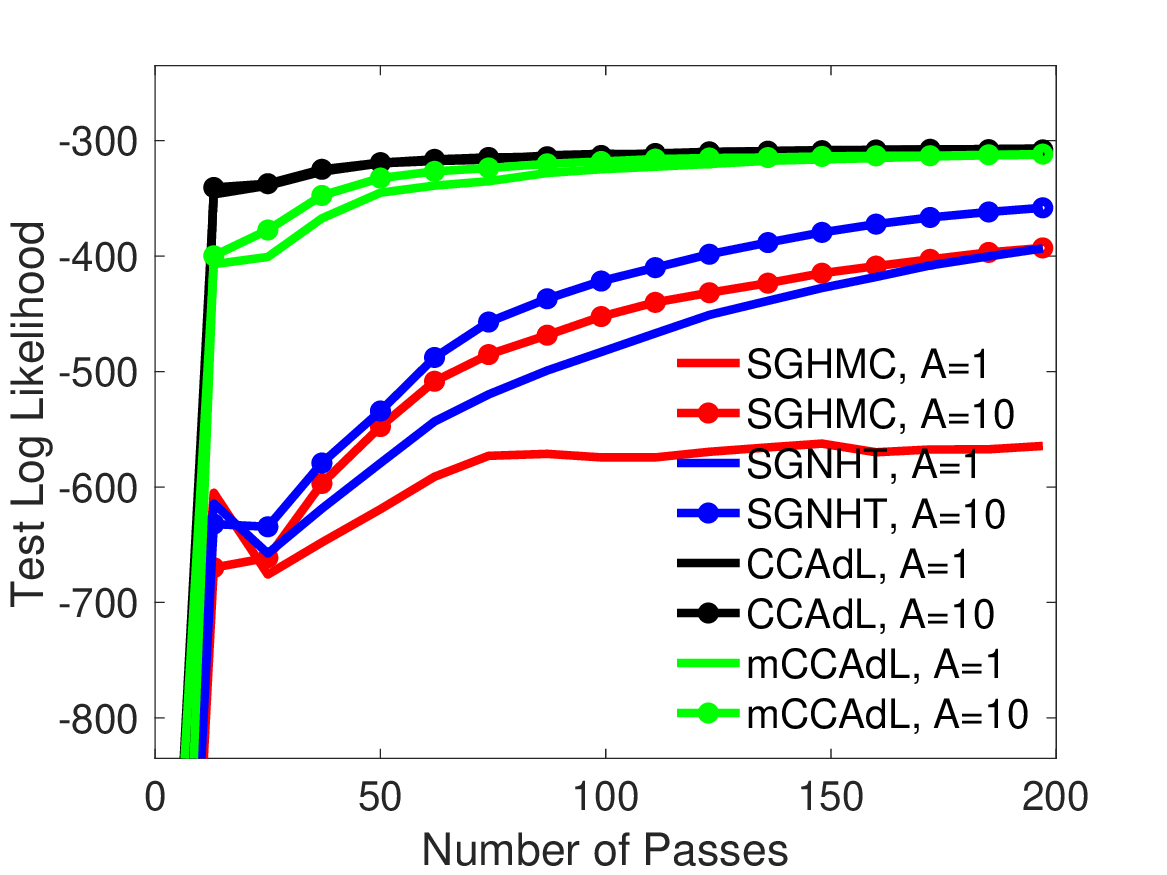} 
\hspace{-4.5mm}
&\includegraphics[width=0.33\linewidth]{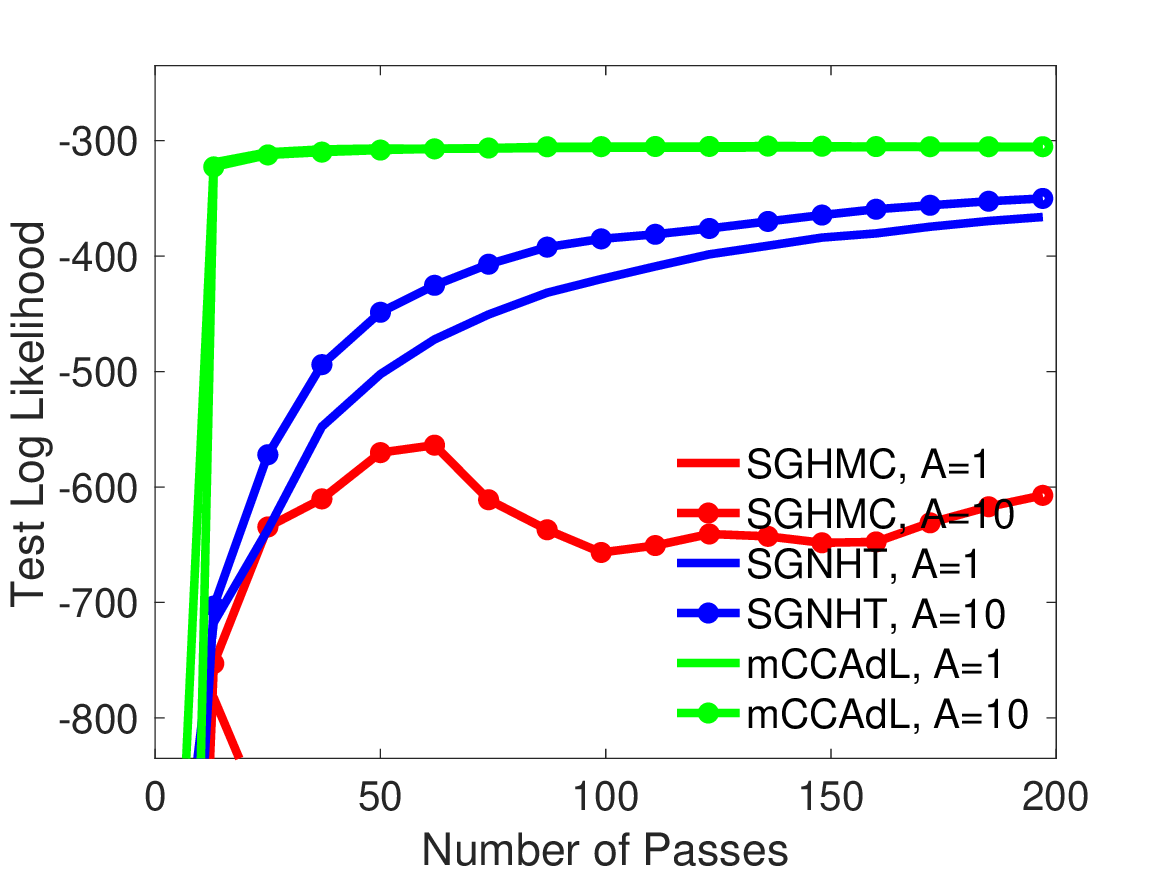} 
\hspace{-4.5mm}
&\includegraphics[width=0.33\linewidth]{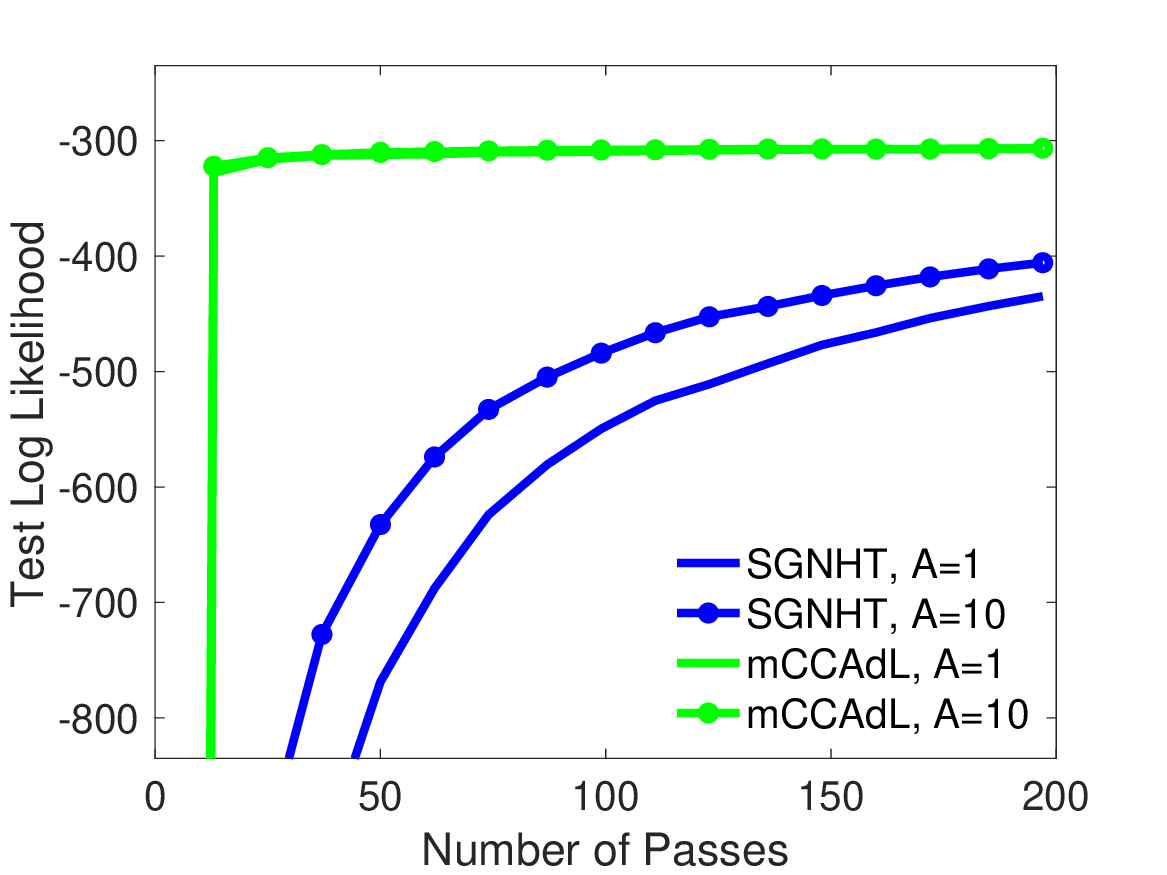}\\

\small{(a) $h=1.2 \times 10^{-4}$} \hspace{-5.5mm} & \small{(b) $h=5 \times 10^{-4}$} \hspace{-2.5mm} & \small{(c) $h=1.2 \times 10^{-3}$}
\end{tabular}
\end{center}
\caption{\small Comparisons of the test log likelihood  of various methods using the posterior mean against the number of passes over the entire dataset in the Bayesian logistic regression on the MNIST dataset of digits 7 and 9 with various values of the stepsize $h$ and effective friction $A$.
}
\label{fig:mnist-pca}
\end{figure}
\begin{figure}[tb]
\begin{center}
\begin{tabular}{ccc}
\includegraphics[width=0.33\linewidth]{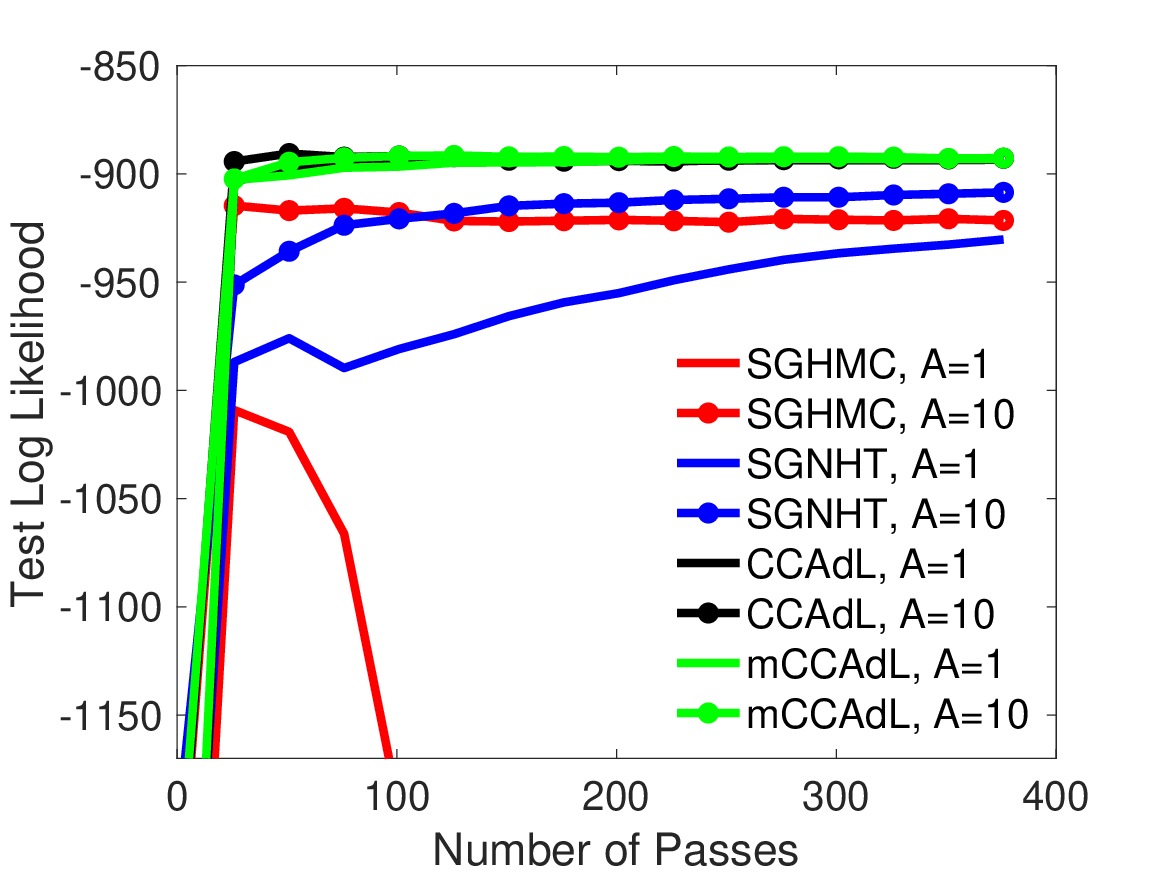} 
\hspace{-4.5mm}
&\includegraphics[width=0.33\linewidth]{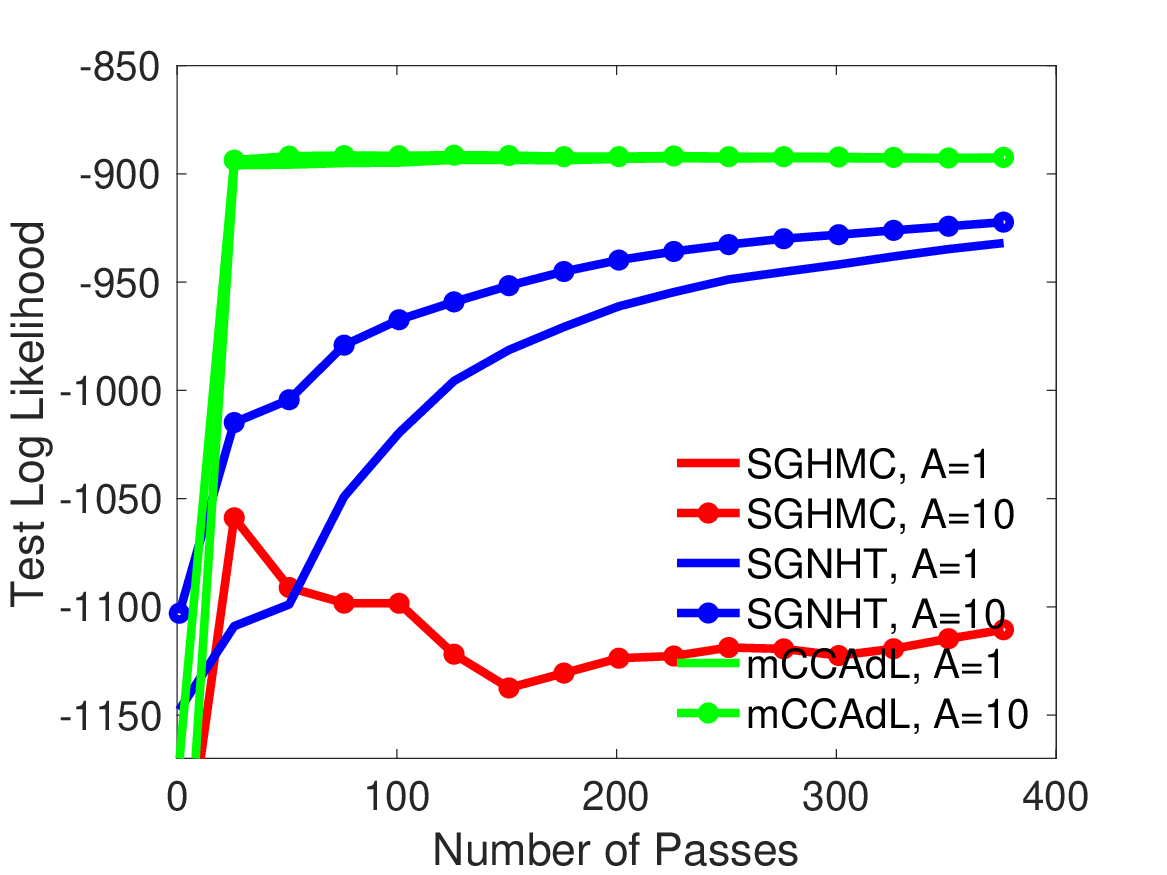} 
\hspace{-4.5mm}
&\includegraphics[width=0.33\linewidth]{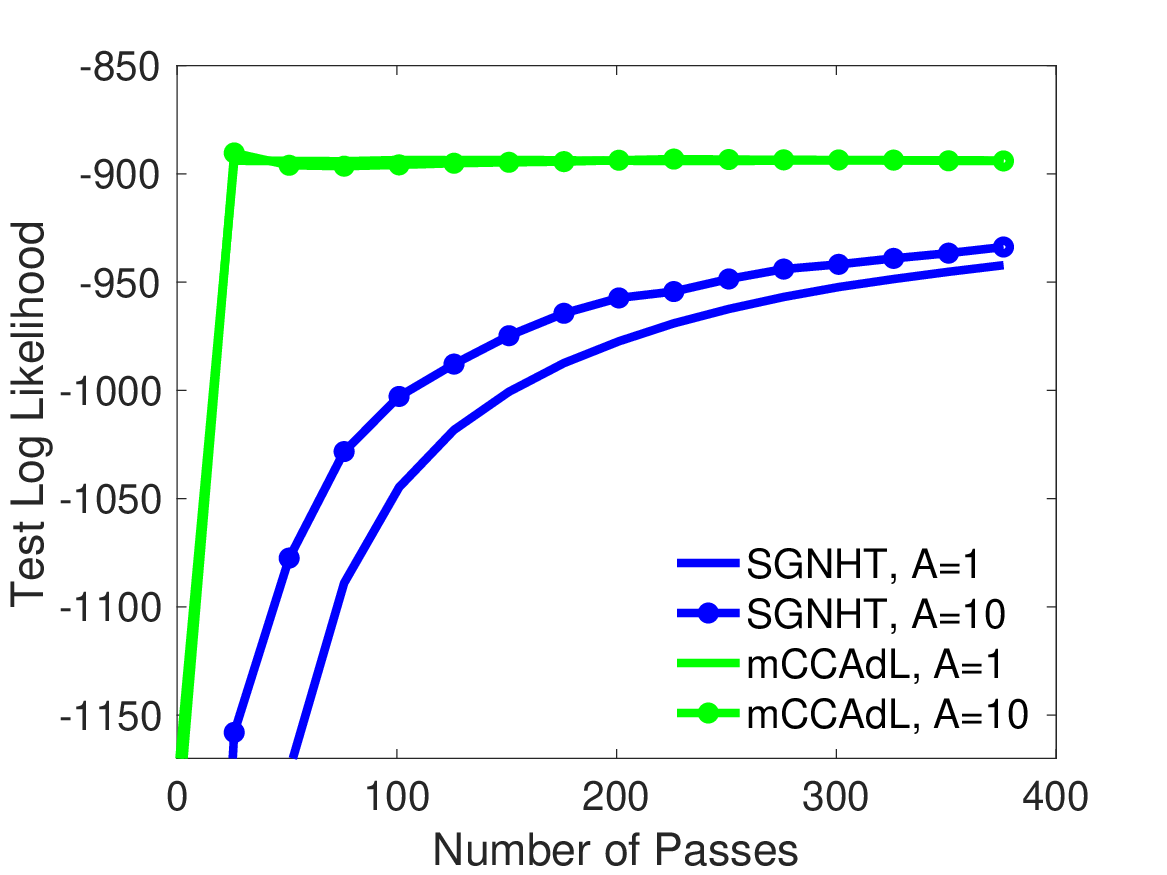}\\
\small{(1a) $h=3.8 \times 10^{-4}$} \hspace{-4.5mm} & \small{(1b) $h=1 \times 10^{-3}$} \hspace{-5.5mm} & \small{(1c)  $h=1.5 \times 10^{-3}$}\\
\includegraphics[width=0.33\linewidth]{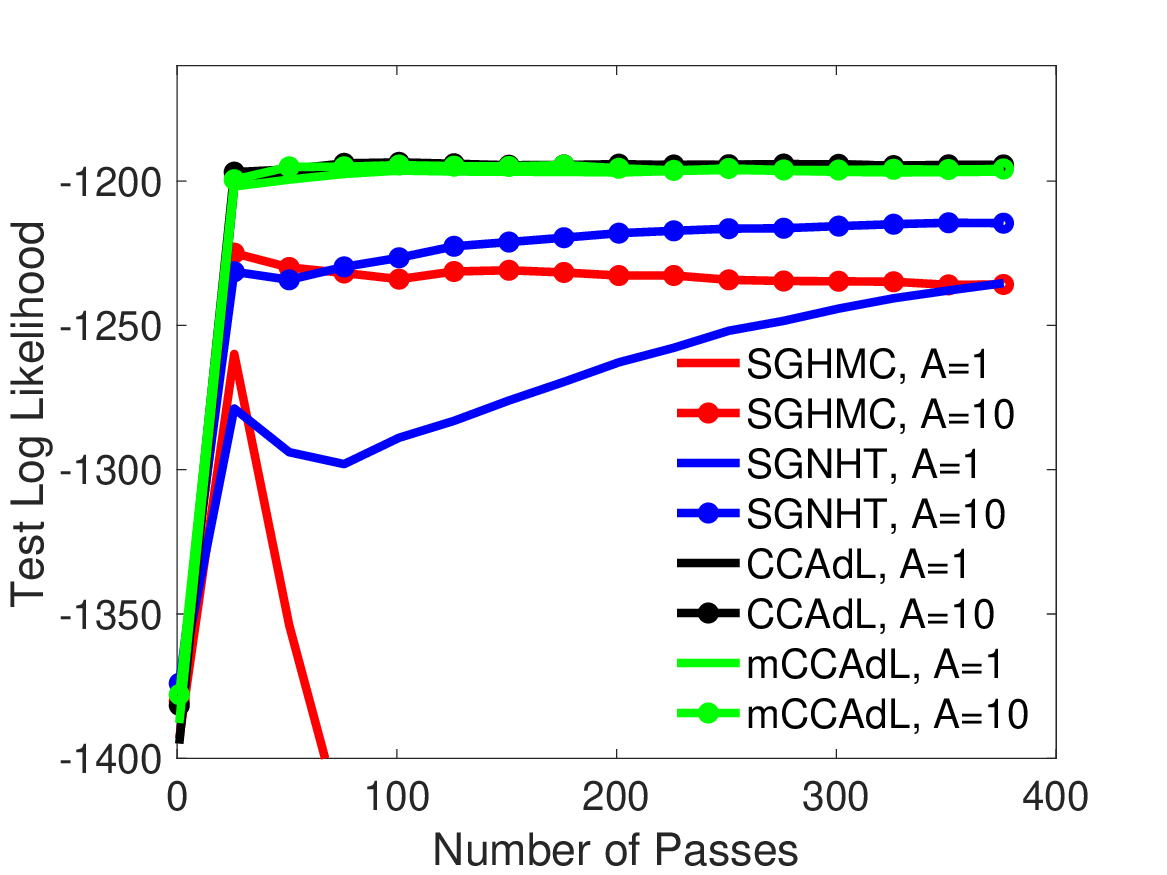} 
\hspace{-4.5mm}
&\includegraphics[width=0.33\linewidth]{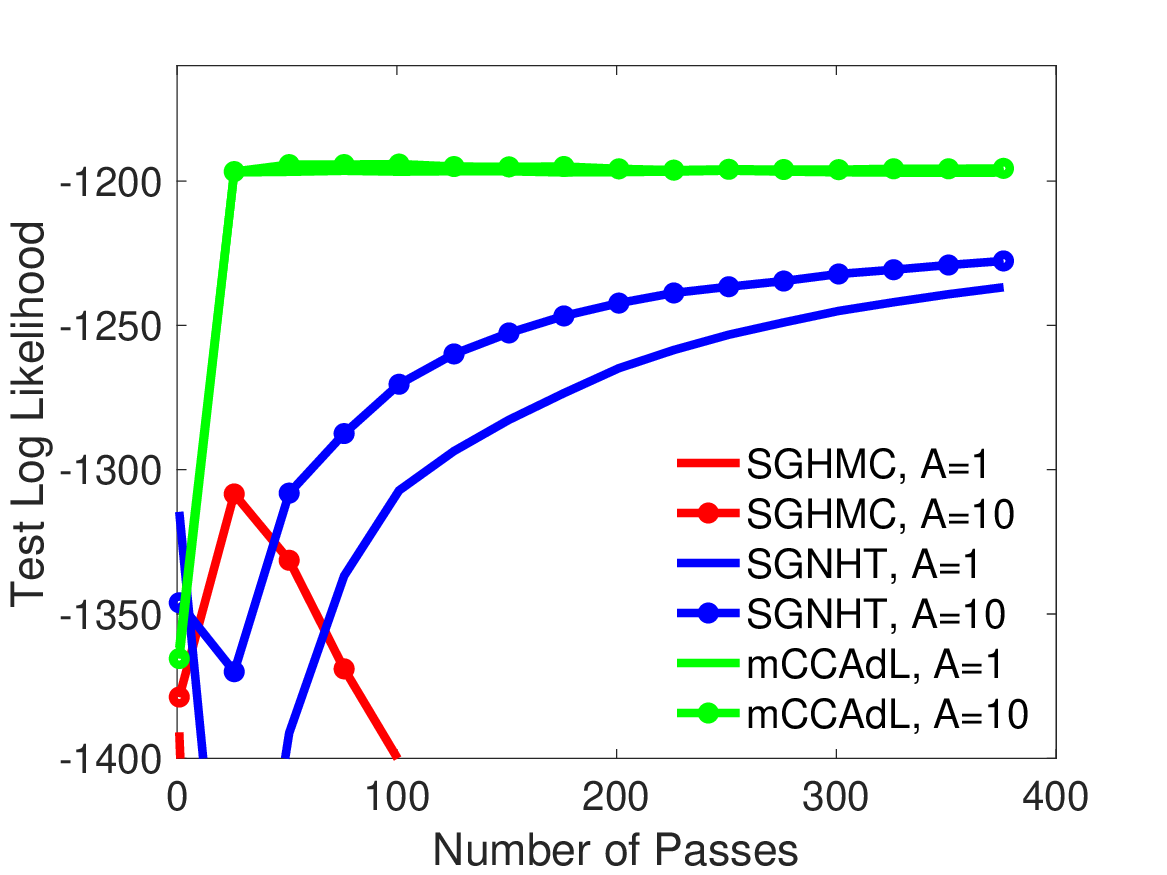} 
\hspace{-4.5mm}
&\includegraphics[width=0.33\linewidth]{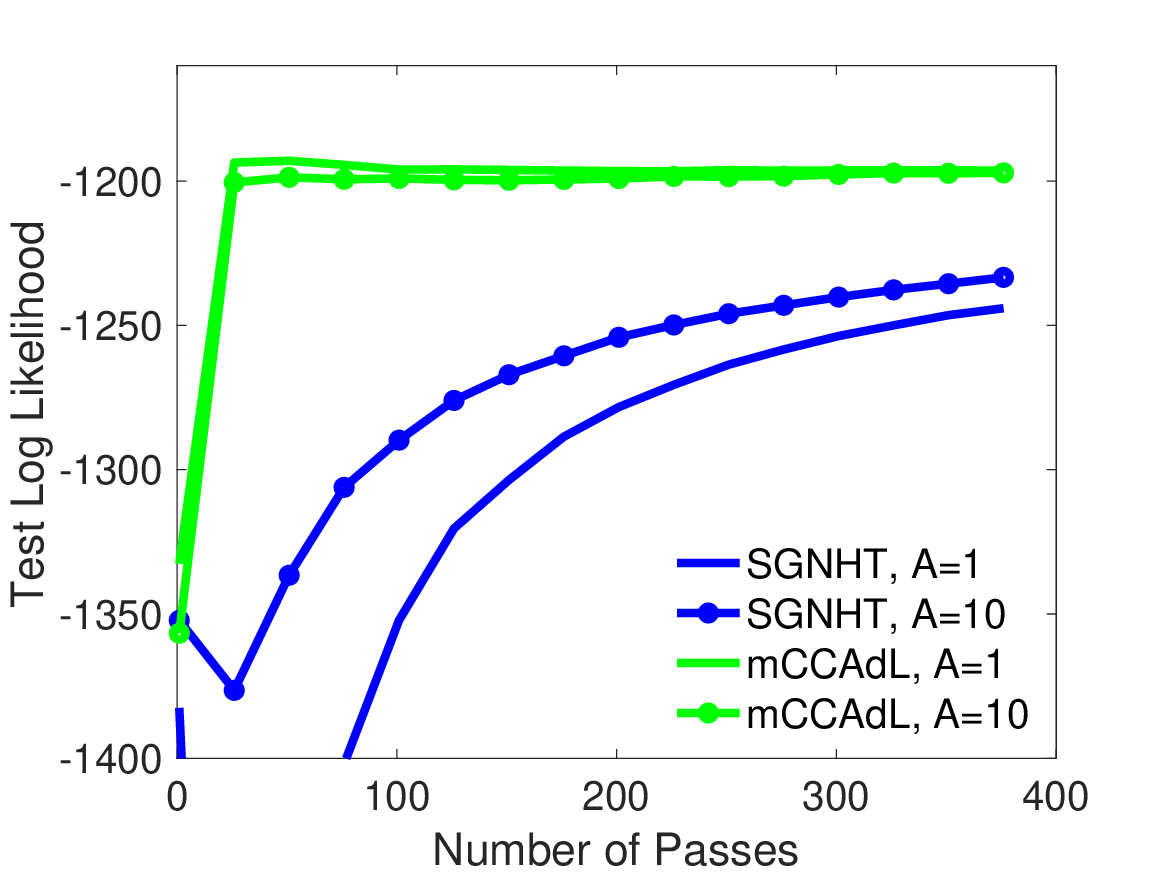}\\
\small{(2a)  $h=3.9 \times 10^{-4}$}  \hspace{-4.5mm} & \small{(2b)  $h=1 \times 10^{-3}$} \hspace{-5.5mm} & \small{(2c)  $h=1.5 \times 10^{-3}$}\\
\includegraphics[width=0.33\linewidth]{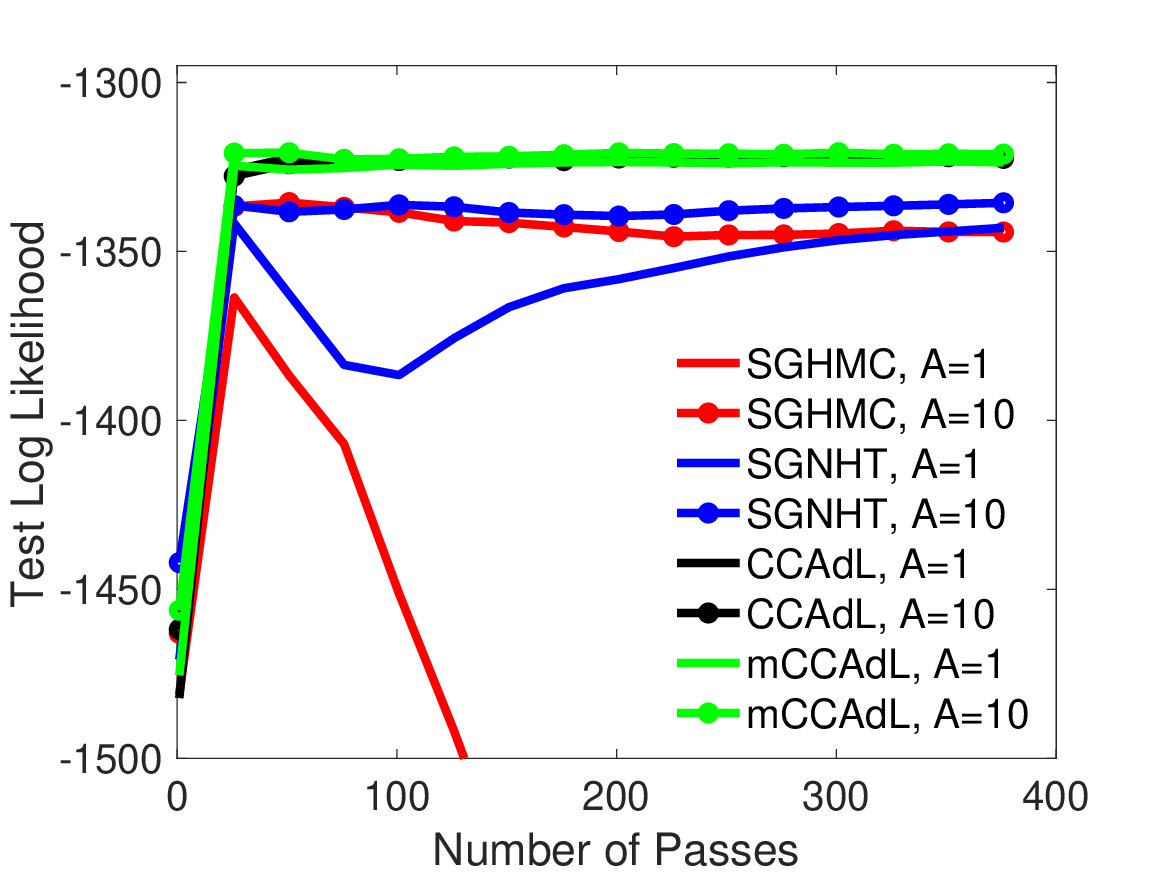} 
\hspace{-4.5mm}
&\includegraphics[width=0.33\linewidth]{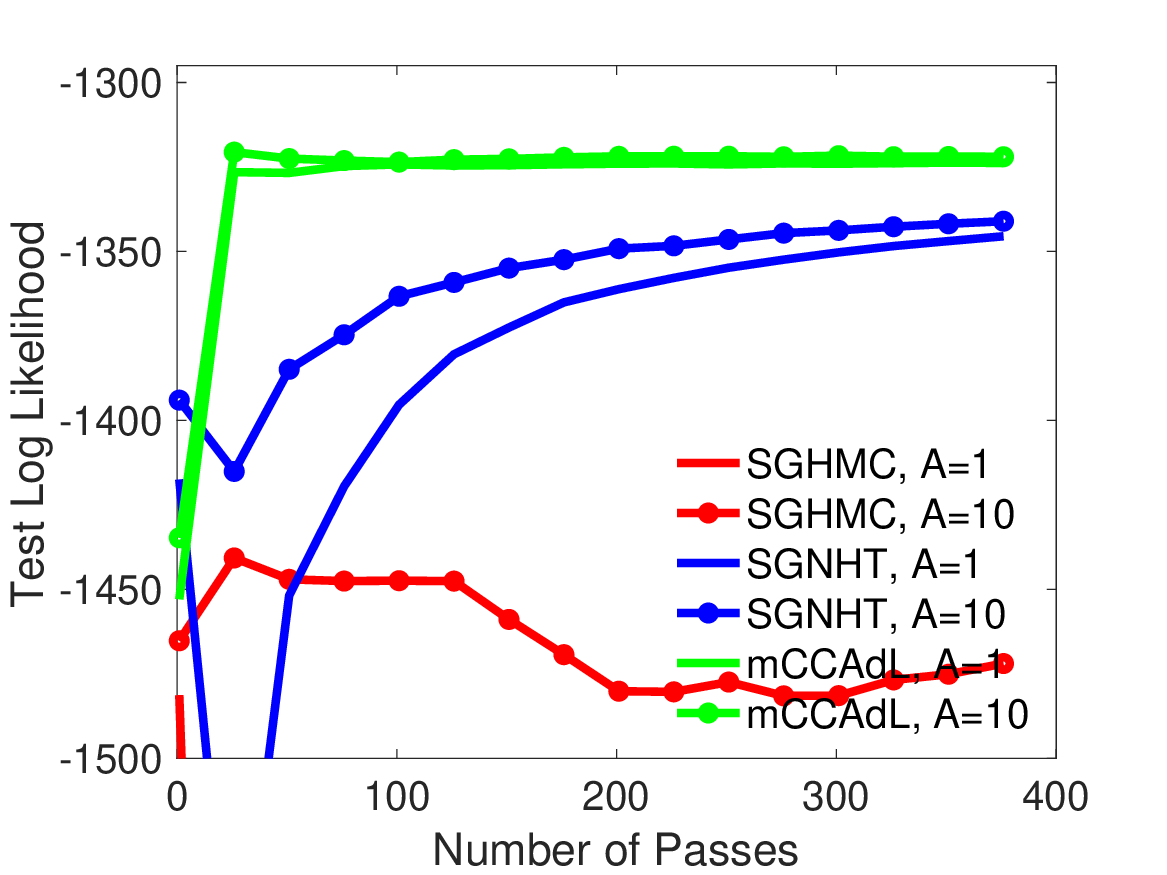}
\hspace{-4.5mm}
&\includegraphics[width=0.33\linewidth]{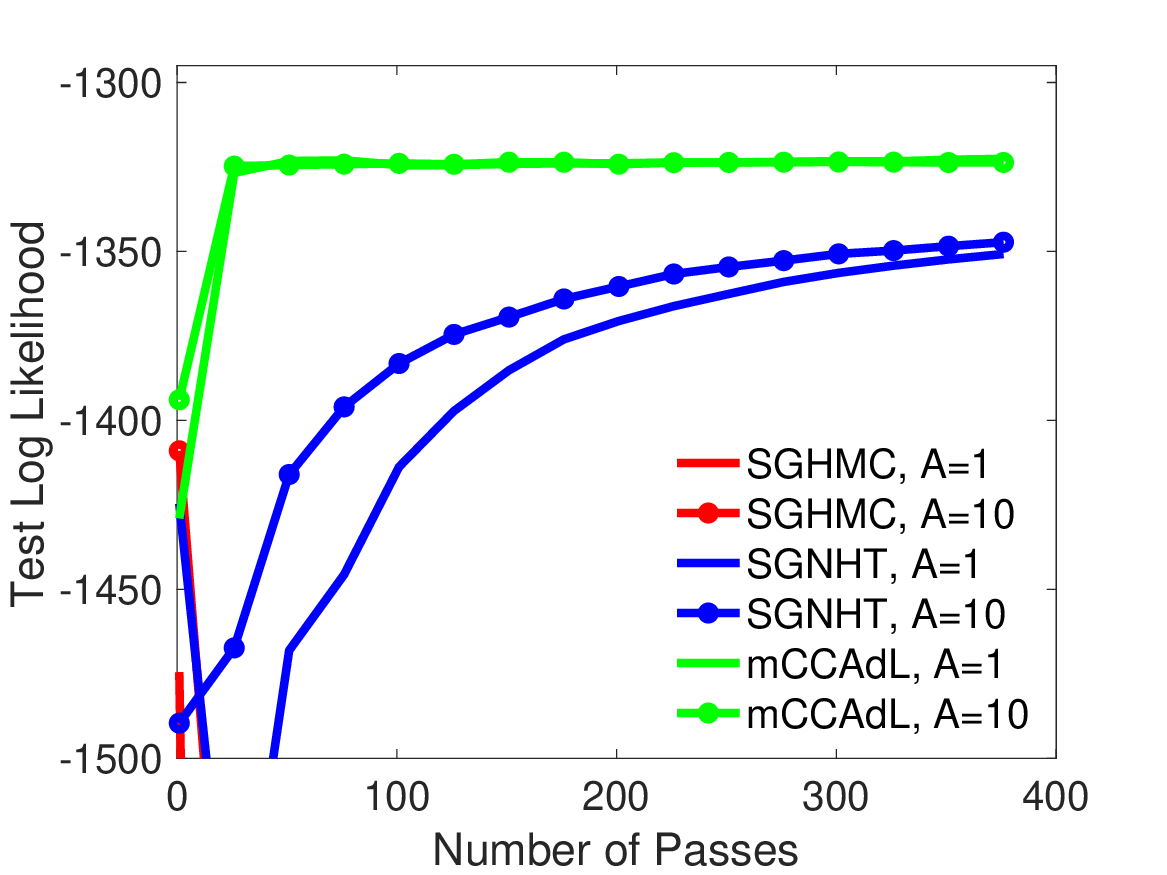}\\
\small{(3a)  $h=3.5 \times 10^{-4}$} \hspace{-5.5mm} & \small{(3b)  $h=1 \times 10^{-3}$} \hspace{-7.5mm} & \small{(3c)  $h=1.5 \times 10^{-3}$}
\end{tabular}
\end{center}
\caption{\small Comparisons of the test log likelihood  of various methods using the posterior mean against the number of passes over the entire dataset in the Bayesian logistic regression on the CIFAR-10 dataset of airplane and automobile (top row), deer and horse (middle row), and cat and dog (bottom row) with various values of the stepsize $h$ and effective friction $A$.}
\label{fig:cifar-10}
\end{figure}

Figure~\ref{fig:mnist-pca} and Figure~\ref{fig:cifar-10} compare the convergence behavior of each method by evaluating the test log likelihood using the posterior mean, plotted against the number of passes over the entire dataset. It is worth mentioning that the largest stepsize used in the original CCAdL article~\cite{Shang2015} was $h = 1 \times 10^{-4}$. In contrast, we used much larger stepsizes in Figure~\ref{fig:mnist-pca} in order to demonstrate the improvement in the stability of the newly proposed mCCAdL method.

When the stepsize is $h = 1.2 \times 10^{-4}$ on the left panel of Figure~\ref{fig:mnist-pca}, we can see that CCAdL converges slightly faster mCCAdL, while clearly outperforming SGHMC and SGNHT: (1) both CCAdL and mCCAdL methods converge much faster than SGHMC and SGNHT, indicating their faster mixing speed and shorter burn-in period; (2) both CCAdL and mCCAdL methods show robustness with different values of the effective friction $A$, whereas SGHMC and SGNHT rely on a relative large value of the effective friction $A$ (especially for SGHMC) that is intended to dominate the gradient noise. When the stepsize was increased, CCAdL quickly became unstable and was therefore not included in further comparisons in Figure~\ref{fig:mnist-pca}. When the stepsize is $h = 5 \times 10^{-4}$ on the middle panel of the figure, while mCCAdL again significantly outperforms SGHMC and SGNHT, the SGHMC method appears to be notably worse, even with a relative large value of the effective friction $A$. When the stepsize was further increased, SGHMC also became unstable and was therefore not included in further comparisons in Figure~\ref{fig:mnist-pca}. When the stepsize is $h = 1.2 \times 10^{-3}$ on the right panel of the figure, the mCCAdL method is remarkably still stable and performs similarly as in the previous two cases. However, in stark contrast, even SGNHT appears to be substantially worse than in the previous two cases. This demonstrates that mCCAdL has substantially improved the stability than the other three alternative methods. Particularly, the largest usable stepsize in mCCAdL (i.e., $h = 1.2 \times 10^{-3}$) is over an order of magnitude larger than that in the original CCAdL thermostat (i.e., $h = 1 \times 10^{-4}$).

The performance of various methods is largely similar in Figure~\ref{fig:cifar-10} for the CIFAR-10 dataset. On the left panel, CCAdL and mCCAdL behave similarly, while clearly outperforming SGHMC and SGNHT. On the middle panel with an increased stepsize, CCAdL became unstable and was therefore not included; mCCAdL again significantly outperforms SGHMC and SGNHT, with the SGHMC method appearing to be notably worse, even with a relatively large value of the effective friction $A$. With further increases in the stepsize on the right panel, SGHMC also became unstable and was therefore not included; mCCAdL remains remarkably stable, with SGNHT performing substantially worse than in the previous two cases.

To complement the above findings based on the test log likelihood, we also evaluate the predictive performance of the algorithms using the log loss, a widely adopted metric for measuring classifier accuracy on a test set $T_*$, for which $|T_*|$ denotes the size of the test set. While mathematically related to the test log likelihood, the log loss evaluates predictive accuracy at the level of individual test points and is particularly sensitive to overconfident misclassifications, thereby offering a complementary perspective on the posterior predictive distribution. In the binary settings considered here, the log loss is given by
\begin{equation}\label{eq:logloss}
  \ell(\thetaB,T_*) 
  = -\frac{1}{|T_*|} \sum_{i_*=1}^{|T_*|} \log \left(\frac{1}{1 + \exp\left(-y_{i_*} \thetaB^{\mathsf{T}} \x_{i_*}\right)} \right) = \frac{1}{|T_*|} \sum_{i_*=1}^{|T_*|} \log \left( 1 + \exp\left(-y_{i_*} \thetaB^{\mathsf{T}} \x_{i_*}\right) \right) \, ,
\end{equation}
where again the labels are assigned as either $y_i = 1$ or $y_i = -1$. It is worth mentioning that~\eqref{eq:likelihood-MNIST} and~\eqref{eq:logloss} are equivalent to the corresponding functions in~\cite{Nemeth2021} where the labels are assigned as either $y_i = 1$ or $y_i = 0$.

\begin{table}[tb]
\centering
\caption{Comparisons of the posterior expected log losses of various methods on the MNIST and CIFAR-10 datasets with various values of the stepsize $h$ and effective friction $A$.}
\label{tab:logloss_results}
\tiny
\setlength{\tabcolsep}{7pt}
\renewcommand{\arraystretch}{1.7}
\begin{tabular}{|c|c|cccc|cccc|}
\hline
\multirow{2}{*}{} & \multirow{2}{*}{$h$} 
& \multicolumn{4}{c|}{$A=1$} 
& \multicolumn{4}{c|}{$A=10$} \\
\cline{3-10}
& & SGHMC & SGNHT & CCAdL & mCCAdL & SGHMC & SGNHT & CCAdL & mCCAdL \\
\hline
\multirow{3}{*}{\shortstack{MNIST \\ (digits 7 and 9)}}
& $1.2 \times 10^{-4}$ & 0.4186 & 0.2626 & 0.1608 & 0.1698 & 0.2619 & 0.2241 & 0.1582 & 0.1642 \\
& $5 \times 10^{-4}$ & 3.3546 & 0.2410 & NaN    & 0.1555 & 0.4583 & 0.2221 & NaN    & 0.1551 \\
& $1.2 \times 10^{-3}$ & Inf & 0.2945   & NaN    & 0.1622 & 1.1128 & 0.2736   & NaN    & 0.1614 \\
\hline
\multirow{3}{*}{\shortstack{CIFAR-10 \\ (airplane--automobile)}}
& $3.8 \times 10^{-4}$ & 2.3260 & 0.5785 & 0.4664 & 0.4671 & 0.5694 & 0.5229 & 0.4655 & 0.4659 \\
& $1   \times 10^{-3}$ & 6.2227 & 0.5836 & NaN & 0.4688 & 0.8750 & 0.5596 & NaN & 0.4677 \\
& $1.5 \times 10^{-3}$ & 9.8335 & 0.6049 & NaN & 0.4725 & 1.1894 & 0.5840 & NaN & 0.4715 \\
\hline
\multirow{3}{*}{\shortstack{CIFAR-10 \\ (deer--horse)}}
& $3.9 \times 10^{-4}$ & 2.0485  & 0.7538 & 0.6165 & 0.6181 & 0.7567 & 0.6908 & 0.6162 & 0.6173 \\
& $1   \times 10^{-3}$ & 9.5790 & 0.7614 & NaN & 0.6206 & 1.1854 & 0.7363 & NaN & 0.6193 \\
& $1.5 \times 10^{-3}$ & 13.0544 & 0.7847 & NaN & 0.6241 & 1.6577 & 0.7552 & NaN & 0.6237 \\
\hline
\multirow{3}{*}{\shortstack{CIFAR-10 \\ (cat--dog)}}
& $3.5 \times 10^{-4}$ & 2.1113 & 0.8332 & 0.6803 & 0.6813 & 0.8329 & 0.7699 & 0.6804 & 0.6801 \\
& $1   \times 10^{-3}$ & 8.8982 & 0.8475 & NaN & 0.6838 & 1.3210 & 0.8082 & NaN & 0.6825 \\
& $1.5 \times 10^{-3}$ & 15.9131 & 0.8765 & NaN & 0.6880 & 1.9174 & 0.8615 & NaN & 0.6880 \\
\hline
\end{tabular}
\end{table}

Table~\ref{tab:logloss_results} reports the mean log losses with the same settings as in Figure~\ref{fig:mnist-pca} and Figure~\ref{fig:cifar-10}. Following~\cite{Nemeth2021}, the mean log loss corresponds to the posterior expected log loss on each test dataset, obtained by first computing the log loss for each posterior sample, and then averaging these values across all posterior samples. Since the predictive distribution is directly determined by the posterior sampling, this metric can also measure the quality of the sampling. These results are consistent with those obtained from the test log likelihood. For the smallest stepsize in each dataset, the mean log losses of CCAdL and mCCAdL are very similar to each other, while clearly outperforming SGHMC and SGNHT. However, as the stepsize increases, CCAdL rapidly becomes unstable, leading to unreliable predictive accuracy and indicating a deterioration in sampling quality, as reflected by an undefined (shown as NaN) mean log loss.  In contrast, mCCAdL remains stable with various values of the stepsize and consistently generates accurate posterior samples together with strong predictive performance, thereby demonstrating superior robustness over the other three alternative methods.


\subsection{Multiclass classification with discriminative restricted Boltzmann machine}
\label{subsec:DRBM}

We also conduct numerical experiments with the discriminative restricted Boltzmann machine~\cite{Larochelle2008}, which serves as a nonlinear classifier with a tractable discriminative objective. The model was trained on two large-scale multiclass datasets from the LIBSVM collection\footnote{\small \url{http://www.csie.ntu.edu.tw/~cjlin/libsvmtools/datasets/multiclass.html}}, namely the \emph{letter} and \emph{SensIT Vehicle (acoustic)} datasets, whose detailed information is presented in Table~\ref{tab:datasets}. Note that the number of hidden units was chosen via cross-validation in order to achieve their best results, leading to 100 hidden units for the \emph{letter} dataset and 20 for the \emph{acoustic} dataset, respectively. 

\begin{table}[tb]
\caption{Detailed information of the datasets used in the discriminative restricted Boltzmann machine.}
\vspace{3mm}
\centering
\begin{tabular}{cccccc}
\toprule
Datasets & training/test set & classes & features & hidden units & total number of parameters $N_\mathrm{d}$ \\
\midrule
\emph{letter} & 10,500/5,000 & 26 & 16 & 100 & 4326 \\
\midrule
\emph{acoustic} & 78,823/19,705 & 3 & 50 & 20 & 1083 \\
\bottomrule
\end{tabular}
\label{tab:datasets}
\end{table}


In contrast to the original CCAdL article~\cite{Shang2015}, which employed a diagonal approximation of the covariance matrix for improving the computational efficiency, we considered the full covariance matrix in this article in order to ensure a fair comparison of various methods. For both datasets, the size of the subset was chosen as $n = 1000$ with replacement. In addition, the initial 20\% of the total number of passes over the dataset were discarded as burn-in, and the remaining samples were collected for prediction. 
\begin{figure}[tb]
\begin{center}
\begin{tabular}{ccc}
\includegraphics[width=0.33\linewidth]{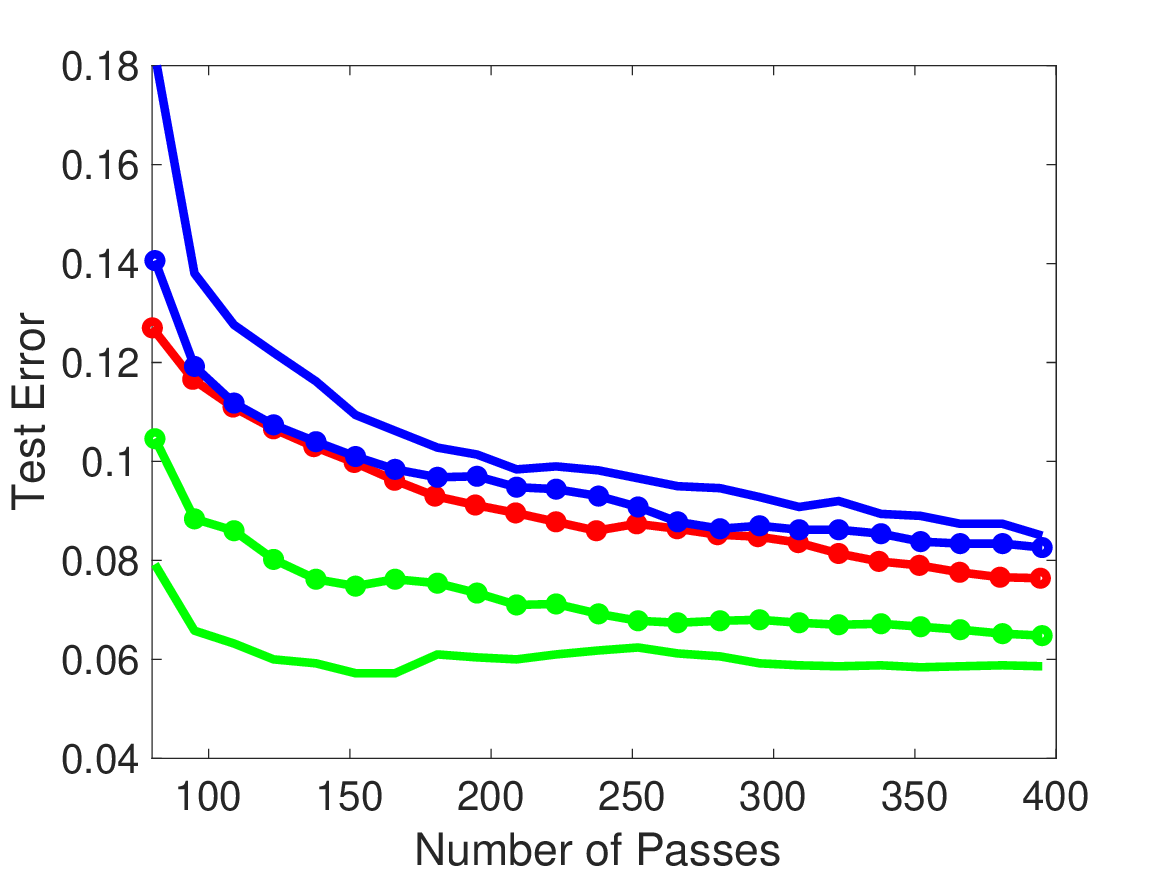}  
\hspace{-4.5mm}
&\includegraphics[width=0.33\linewidth]{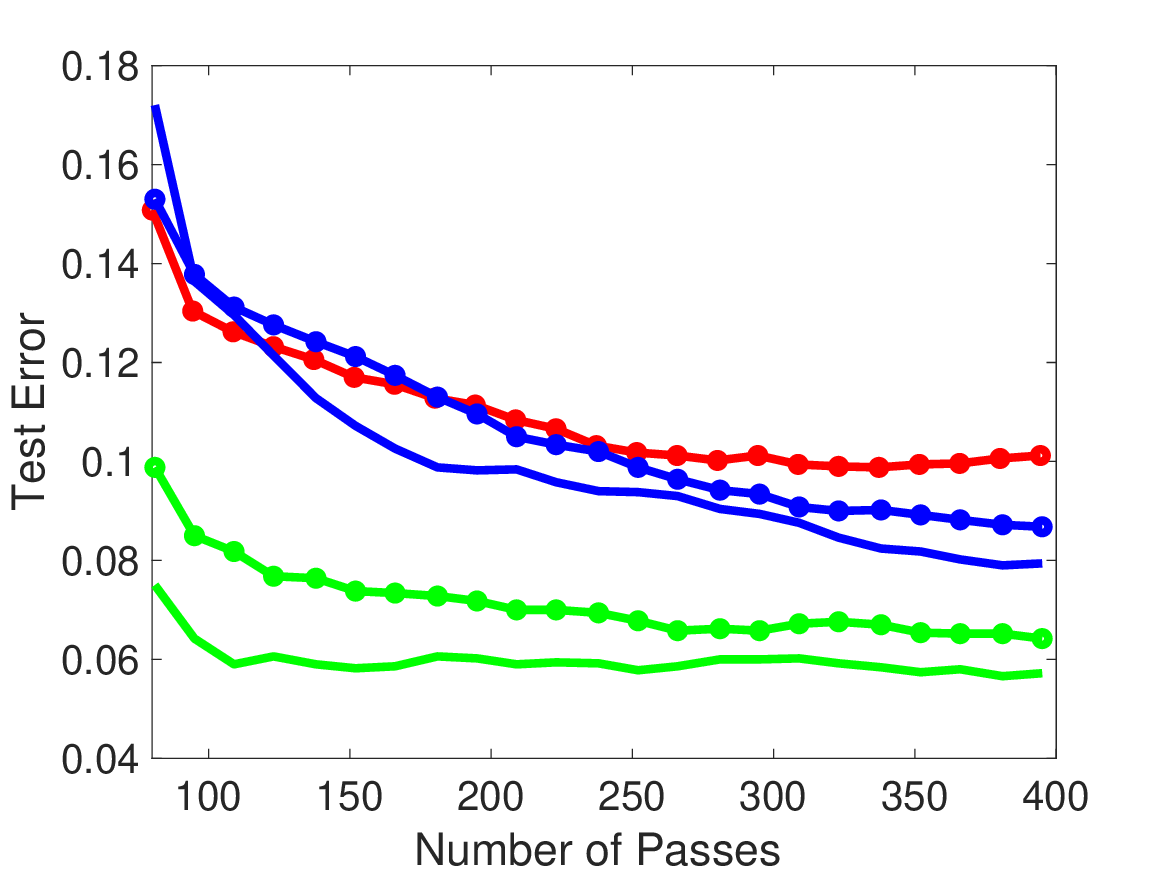} 
\hspace{-4.5mm}
&\includegraphics[width=0.33\linewidth]{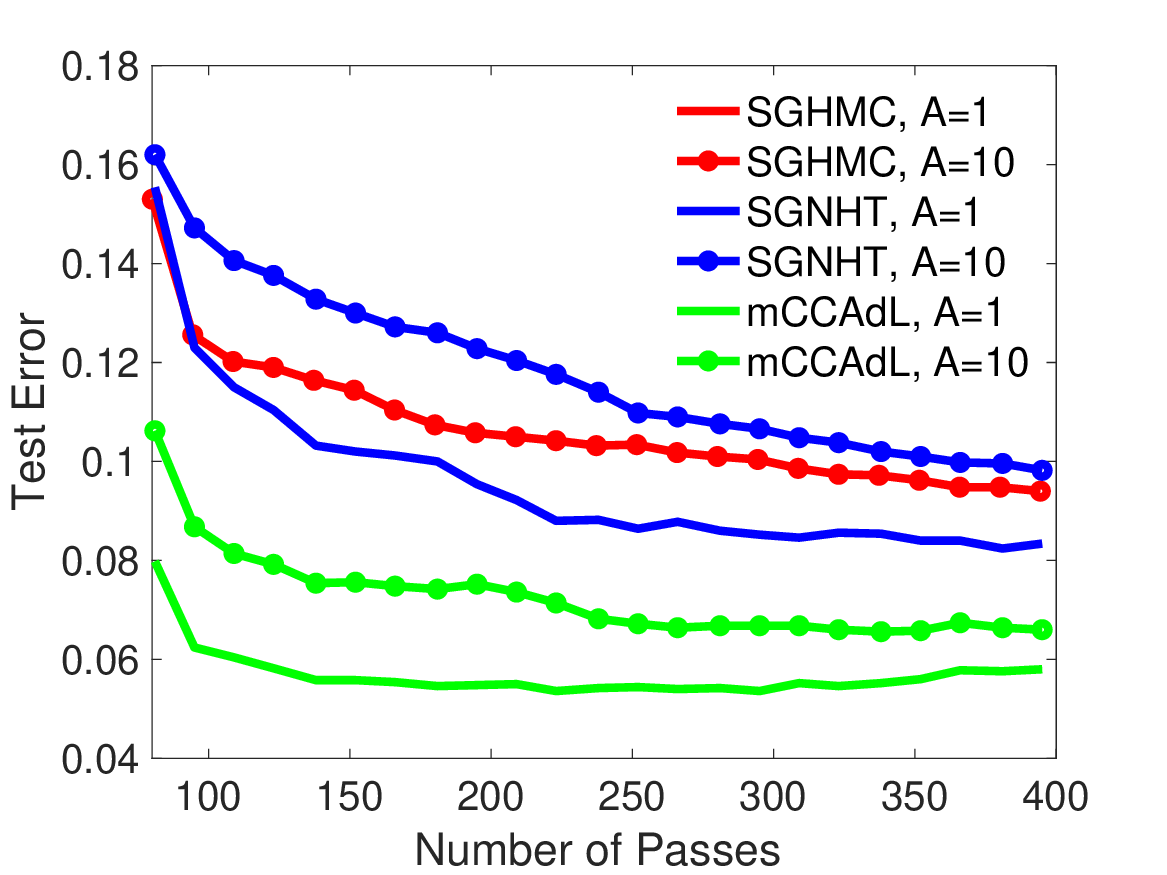}
\\
\small{(1a) \emph{letter}, $h=2 \times 10^{-2}$}  \hspace{-4.5mm} & \small{(1b) \emph{letter}, $h=2.5 \times 10^{-2}$} \hspace{-5.5mm} & \small{(1c) \emph{letter}, $h=3 \times 10^{-2}$}\\
\includegraphics[width=0.33\linewidth]{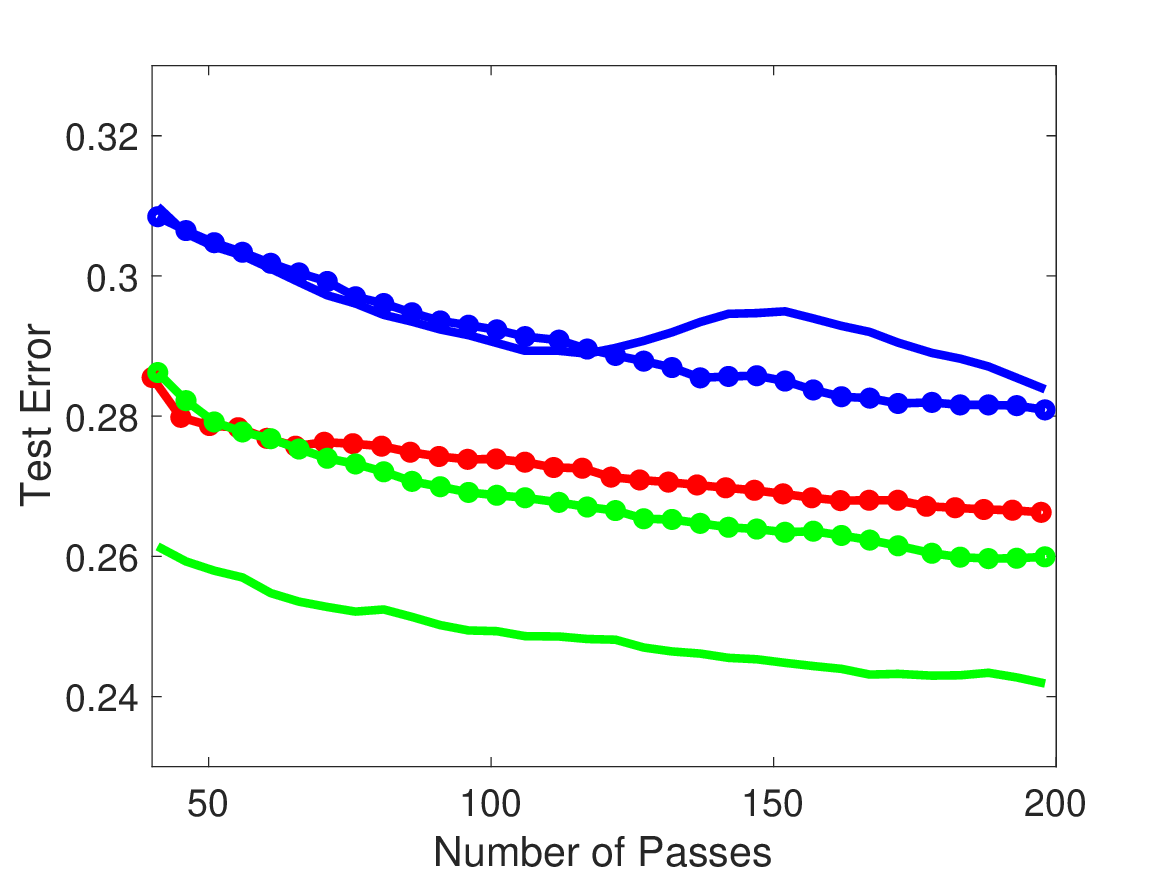} 
\hspace{-4.5mm}
&\includegraphics[width=0.33\linewidth]{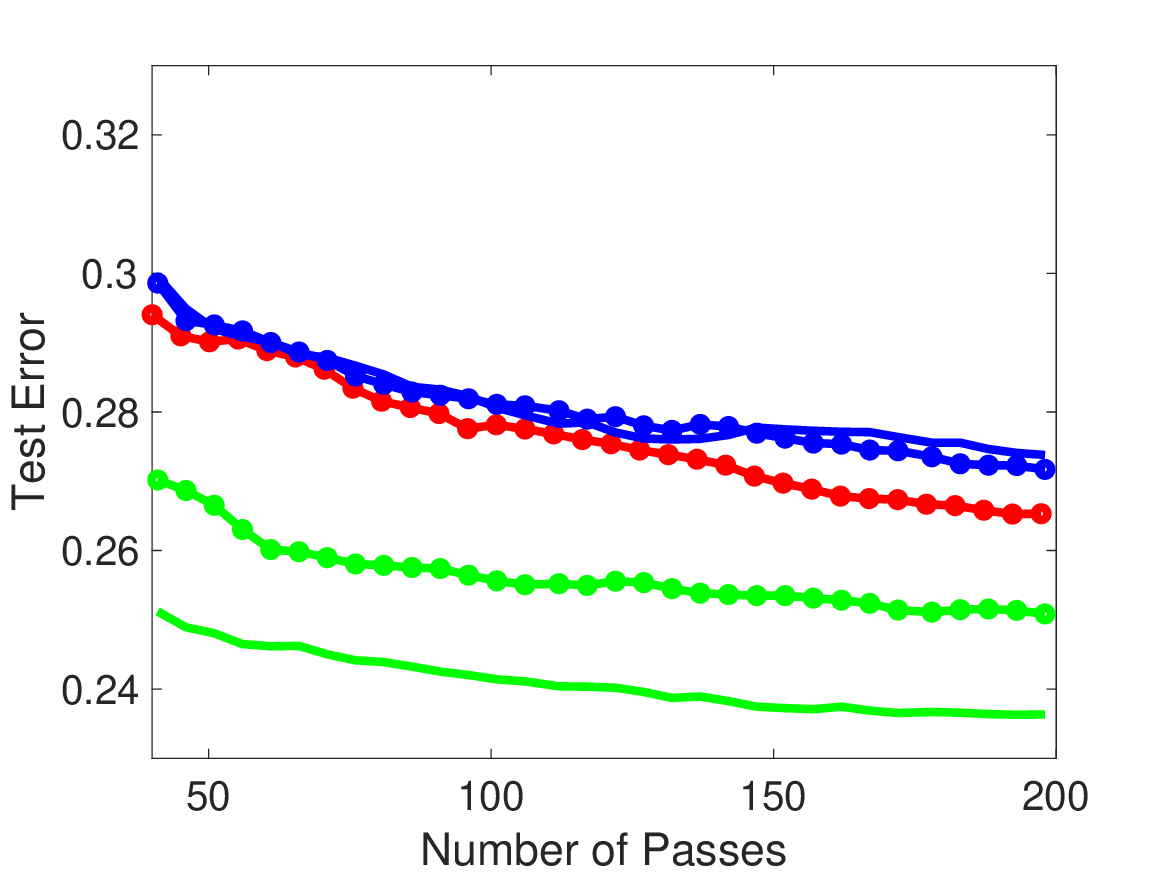}
\hspace{-4.5mm}
&\includegraphics[width=0.33\linewidth]{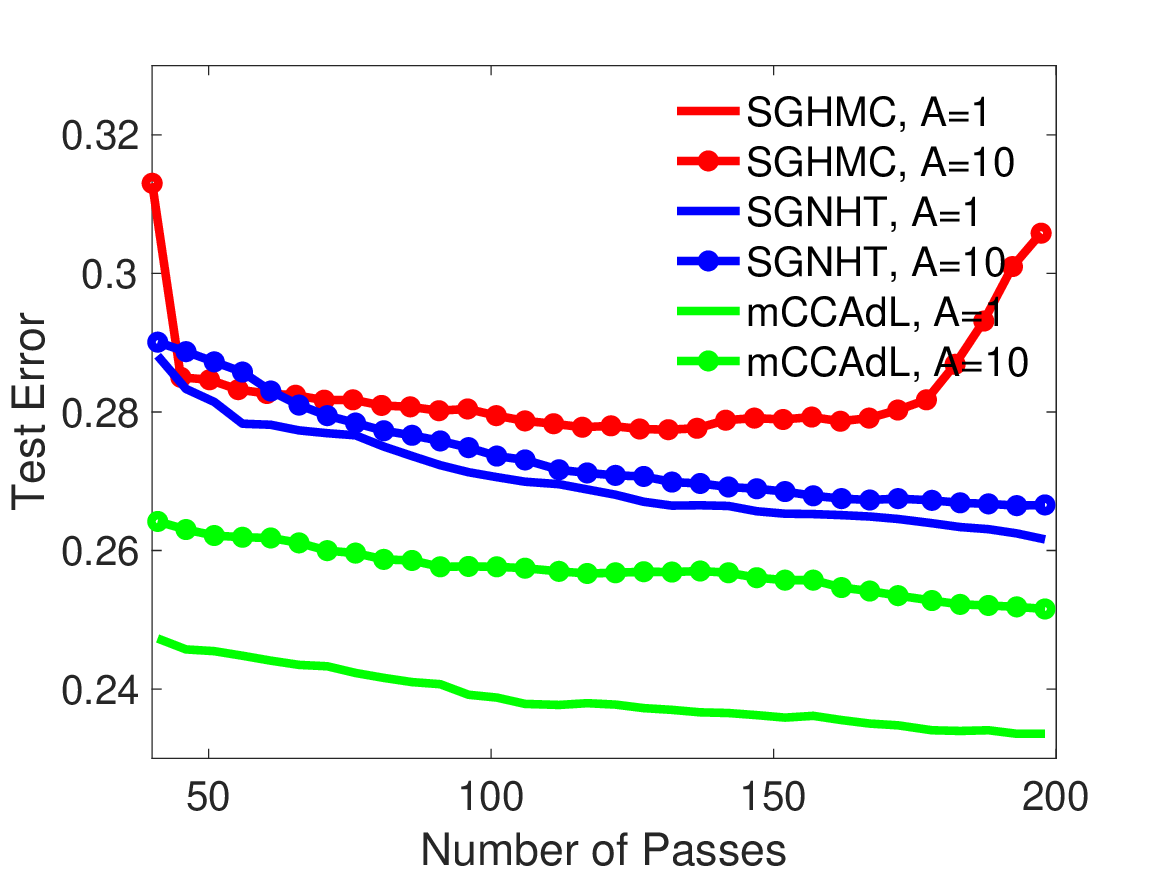}\\
\small{(2a) \emph{acoustic}, $h=1.5 \times 10^{-3}$} \hspace{-5.5mm} & \small{(2b) \emph{acoustic}, $h=2.5 \times 10^{-3}$} \hspace{-7.5mm} & \small{(2c) \emph{acoustic}, $h=3.5 \times 10^{-3}$}
\end{tabular}
\end{center}
\caption{\small Comparisons of the test error rates of various methods using the posterior mean against the number of passes over the entire dataset on datasets \emph{letter} (top row) and \emph{acoustic} (bottom row) with various values of the stepsize $h$ and effective friction $A$ indicated.}
\label{fig:testerror_DRBM}
\end{figure}

Figure~\ref{fig:testerror_DRBM} compares various methods in terms of the test error rates, computed with the posterior mean and plotted against the number of passes over the entire dataset. We again used (much) larger stepsizes in Figure~\ref{fig:testerror_DRBM} than those used in the original CCAdL article~\cite{Shang2015}. CCAdL became unstable for such large stepsizes and was therefore not included. We can clearly see that mCCAdL, particularly with a small value of the effective friction $A$, substantially outperforms SGHMC and SGNHT. Particularly, as we increase the stepsize on the top row of Figure~\ref{fig:testerror_DRBM}, the test error rates of the mCCAdL method remain largely similar, in stark contrast both SGHMC and SGNHT appear to be worse and worse. If the stepsize was large enough, both SGHMC and SGNHT became unstable, while mCCAdL was remarkably still stable with similar accuracy. We also note that SGHMC performs well only when the effective friction $A$ is chosen to be large. However, such a setting leads to a pronounced random-walk behaviour, thereby reducing the rate of convergence.

\begin{table}[tb]
\centering
\caption{Comparisons of the posterior expected log losses of various methods on the \emph{letter} and \emph{acoustic} datasets with various values of the stepsize $h$ and effective friction $A$.}
\label{tab:logloss_results_drbm}
\tiny
\setlength{\tabcolsep}{9.5pt}
\renewcommand{\arraystretch}{1.7}
\begin{tabular}{|c|c|cccc|cccc|}
\hline
\multirow{2}{*}{} & \multirow{2}{*}{$h$} 
& \multicolumn{4}{c|}{$A=1$} 
& \multicolumn{4}{c|}{$A=10$} \\
\cline{3-10}
& & SGHMC & SGNHT & CCAdL & mCCAdL & SGHMC & SGNHT & CCAdL & mCCAdL \\
\hline
\multirow{3}{*}{\shortstack{\emph{letter}}}
& $2   \times 10^{-2}$ & 1.5244 & 0.4269 & NaN & 0.4320 & 0.3299 & 0.3029 & NaN & 0.2656 \\
& $2.5 \times 10^{-2}$ & 2.4337 & 0.3908 & NaN & 0.4692 & 0.3740 & 0.3375 & NaN & 0.2764 \\
& $3   \times 10^{-2}$ & 2.5606 & 0.4305 & NaN & 0.4572 & 0.3735 & 0.3467 & NaN & 0.2770 \\
\hline
\multirow{3}{*}{\shortstack{\emph{acoustic}}}
& $1.5 \times 10^{-3}$ & 0.8727 & 0.6568 & NaN & 0.5787 & 0.6338 & 0.6594 & NaN & 0.6137 \\
& $2.5 \times 10^{-3}$ & 0.9888 & 0.6415 & NaN & 0.5656 & 0.6338 & 0.6445 & NaN & 0.5949 \\
& $3.5 \times 10^{-3}$ & 0.9757 & 0.6279 & NaN & 0.5615 & 0.6543 & 0.6401 & NaN & 0.5933 \\
\hline
\end{tabular}
\end{table}

In addition to the comparisons of the test error rates, we also computed the mean log losses to further assess the performance of various methods, as discussed in Section~\ref{subsec:binaryclass}. 
Table~\ref{tab:logloss_results_drbm} shows the mean log losses on the \emph{letter} and \emph{acoustic} datasets with the same settings as in Figure~\ref{fig:testerror_DRBM}. CCAdL again became unstable for such large stepsizes as in Figure~\ref{fig:testerror_DRBM}, resulting in NaN in the table. For the \emph{letter} dataset, when the effective friction is relatively large as $A=10$, mCCAdL achieves the smallest mean log loss and clearly outperforms both SGHMC and SGNHT. When the effective friction is relatively small as $A=1$, SGNHT appears to be slightly better than mCCAdL, whereas SGHMC performs substantially worse and only becomes competitive when $A$ becomes relatively large. For the \emph{acoustic} dataset, mCCAdL consistently has the smallest mean log loss with all stepsizes compared and for both $A=1$ and $A=10$, while SGHMC and SGNHT produce noticeably larger mean log losses. Therefore, in both datasets, these results are largely consistent with the test error rates reported earlier and highlight that mCCAdL consistently provides more accurate and stable predictive performance with various values of the stepsize.

\section{Conclusions}
\label{sec:Conclusions}

In this article, we have proposed a modified version of the covariance-controlled adaptive Langevin (CCAdL) thermostat originally proposed in~\cite{Shang2015}. While CCAdL relies on a moving average to estimate the covariance matrix of the noisy force generated from subsampling, in the newly proposed mCCAdL method we have used the scaling part of the scaling and squaring method together with a truncated Taylor series approximation to the exponential to numerically approximate the exact solution to the subsystem involving the additional term (i.e., the C part). We have further proposed a symmetric splitting method for mCCAdL. We have demonstrated in large-scale machine learning applications that the newly proposed mCCAdL method achieves a substantial improvement in numerical stability (measured by the largest usable stepsize) over the original CCAdL method, while significantly outperforming popular alternative stochastic gradient methods in terms of accuracy (measured by the test error). The methodologies presented in this article could be of use in more general settings of large-scale Bayesian sampling and optimisation, which we leave for future work.

\section*{Acknowledgments}

The authors thank Gabriel Stoltz and Benedict Leimkuhler
for their valuable suggestions and comments.

\bibliographystyle{is-abbrv}

\bibliography{refs_siam}

\end{document}